\documentclass{article}
\usepackage[utf8]{inputenc} 
\usepackage[T1]{fontenc}    
\usepackage[svgnames,usenames,dvipsnames,table]{xcolor}         
\usepackage[colorlinks=true, citecolor=Navy, breaklinks=true]{hyperref}       
\usepackage{url}            
\usepackage{booktabs}       
\usepackage{amsfonts}       
\usepackage{nicefrac}       
\usepackage{microtype}

\usepackage{algorithm}
\usepackage{algorithmic}

     \PassOptionsToPackage{numbers, compress}{natbib}

\usepackage{cereb}
\usepackage{environ}

\usepackage{amsmath}
\usepackage{amssymb}
\usepackage{mathtools}
\usepackage{amsthm}

\usepackage[capitalize,noabbrev]{cleveref}

\usepackage{wrapfig}
\usepackage{graphicx}
\usepackage{subcaption}
\usepackage{multirow}
\usepackage{tcolorbox}
\usepackage{paralist}

\usepackage{makecell}
\usepackage[table]{xcolor}
\usepackage{colortbl}
\usepackage{tabularx}
\usepackage{siunitx}
\usepackage{xfp} 
\usepackage{refcount}
\usepackage{float}
\usepackage[inkscapelatex=false]{svg}
\usepackage{glossaries}
\makeglossaries

\theoremstyle{plain}

\theoremstyle{definition}

\newtheorem{des}{Desideratum}

\theoremstyle{remark}

\usepackage[textsize=tiny]{todonotes}

\usepackage{shortbold}

\usepackage{amsmath,amsfonts,bm,amssymb}

\def\eqref#1{equation~\ref{#1}}

\def\1{\bm{1}}

\DeclareMathAlphabet{\mathsfit}{\encodingdefault}{\sfdefault}{m}{sl}
\SetMathAlphabet{\mathsfit}{bold}{\encodingdefault}{\sfdefault}{bx}{n}

\usepackage{enumitem}
\setlist[itemize]{left=0.5em}
\setlist[enumerate]{left=0.5em}

\crefname{figure}{Fig.}{Figs.}
\Crefname{figure}{Fig.}{Figs.}
\crefname{equation}{Eq.}{Eqs.}
\Crefname{equation}{Eq.}{Eqs.}
\crefname{section}{Sec.}{Secs.}
\Crefname{section}{Sec.}{Secs.}
\crefname{subsection}{Sec.}{Secs.}
\Crefname{subsection}{Sec.}{Secs.}
\crefname{subsubsection}{Sec.}{Secs.}
\Crefname{subsubsection}{Sec.}{Secs.}
\crefname{theorem}{Theorem}{Theorems}
\Crefname{theorem}{Theorem}{Theorems}
\crefname{lemma}{Lemma}{Lemmas}
\Crefname{lemma}{Lemma}{Lemmas}
\crefname{proposition}{Proposition}{Propositions}
\Crefname{proposition}{Proposition}{Propositions}
\crefname{corollary}{Corollary}{Corollaries}
\Crefname{corollary}{Corollary}{Corollaries}
\crefname{definition}{Definition}{Definitions}
\Crefname{definition}{Definition}{Definitions}
\crefname{assumption}{Assumption}{Assumptions}
\Crefname{assumption}{Assumption}{Assumptions}
\crefname{remark}{Remark}{Remarks}
\Crefname{remark}{Remark}{Remarks}
\crefname{example}{Example}{Examples}
\Crefname{example}{Example}{Examples}

\makeatletter
\@ifundefined{ack}{
  \NewEnviron{ack}{
    \section*{Acknowledgments and Disclosure of Funding}
    \BODY
  }
}{}
\makeatother

\newcounter{fcounter}
\newcommand\finding[1]{
        \refstepcounter{fcounter}\vspace{2pt}
        \begin{tcolorbox}[colback=yellow!10!white,colframe=yellow!80!black,boxsep=1pt,left=2pt,right=2pt,top=1pt,bottom=1pt]\noindent{\textbf{\sffamily Finding \arabic{fcounter}}: \sffamily #1}
        \end{tcolorbox}\vspace{0pt}
}
\crefname{fcounter}{Finding}{Findings}

\newcounter{kcounter}
\newcommand\takeaway[1]{
        \refstepcounter{kcounter}\vspace{2pt}
        \begin{tcolorbox}[colback=green!10!white,colframe=green!80!black,boxsep=1pt,left=2pt,right=2pt,top=1pt,bottom=1pt]\noindent{\textbf{\sffamily Key takeaway \arabic{kcounter}}: \sffamily #1}
        \end{tcolorbox}\vspace{0pt}
}
\crefname{kcounter}{Takeaway}{Takeaways}

\renewcommand{\cite}[1]{\PackageError{MyPackage}{Do not use \string\cite\space with natbib. Use \string\citet\space or \string\citep}{See the natbib package documentation for explanation.}}

\makeatletter
\AtBeginDocument{%
  \@ifpackageloaded{cleveref}{%

    \providecommand\cref@appendix@setup{%
      \crefname{appendix}{Appendix}{Appendices}%
      \Crefname{appendix}{Appendix}{Appendices}%
    }%

    \AddToHook{cmd/appendix/before}{%
      \cref@appendix@setup
      \@ifundefined{chapter}{%
        \crefalias{section}{appendix}%
      }{%
        \crefalias{chapter}{appendix}%
      }%
      \crefalias{subsection}{appendix}%
      \crefalias{subsubsection}{appendix}%
    }%

  }{}
}
\makeatother

\makeatletter
\newcommand{\footnotehyper}[1]{%
  \footnote{\footnotehyper@parse #1\footnotehyper@stop}%
}
\def\footnotehyper@parse\label#1#2\footnotehyper@stop{%
  \hypertarget{#1}{}%
  \label{#1}%
  #2%
}
\makeatother

\newcommand{\crown}{\includegraphics[height=1em]{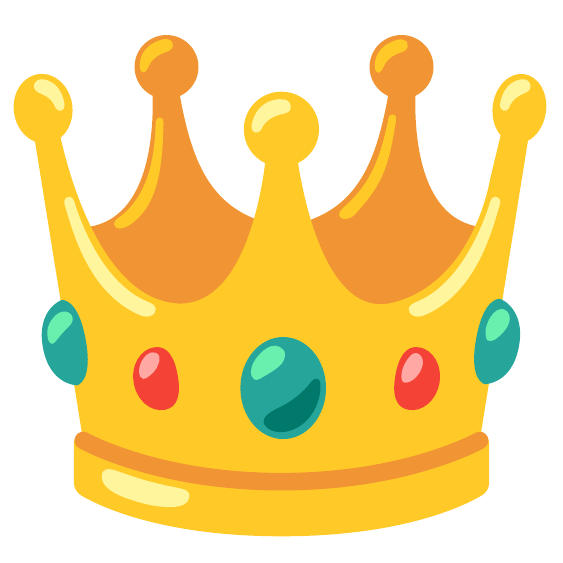}}

\title{Don't Drop Dropout: Optimizing Layer Sparsity for Efficient LLM Training and Inference}
\author{
    Mostafa Elhoushi\textsuperscript{$\dagger$}, Alex Pretko\textsuperscript{$\ddagger$}, Nolan Dey\textsuperscript{$\dagger$}, Bin Claire Zhang\textsuperscript{$\dagger$}, Gavia Gray\textsuperscript{$\dagger$}, Gurpreet Gosal\textsuperscript{$\dagger$}, Abdulrahman Mahmoud\textsuperscript{$\ddagger$}, Shane Bergsma\textsuperscript{$\dagger$}, Joel Hestness\textsuperscript{$\dagger$} \\
    \textsuperscript{$\dagger$}Cerebras Systems, \textsuperscript{$\ddagger$}MBZUAI \\
    \texttt{m.elhoushi@ieee.org, joel@cerebras.net}
}
\date{August 2026}

\begin{document}

\maketitle

\begin{abstract}
    Layer dropout (a.k.a.\ stochastic depth) has been shown to enable faster training, higher accuracy, and robustness to zero-shot layer pruning in both language and vision transformers. However, as models and datasets have scaled, dropout---particularly layer dropout---has largely disappeared from large language models (LLMs) pre-training recipes. While some prior work has reported that dropout can degrade accuracy, no comprehensive study has quantified, let alone mitigated, this effect. In this study, we show that layer dropout \emph{should} be used in state-of-the-art LLM training, establishing best practices and scaling analysis for both training and post-training benefits. Concretely, with optimal layer distribution, time schedule, and optimizer hyperparameters, we observe that \textit{at the same training FLOPs layer dropout leads to lower loss}. For a given number of training steps, LLMs can achieve lower or similar validation loss while saving upto 25\% of training FLOPs. Moreover, layer dropout enables significant post-training optimizations, such as early exit, intermediate-layer skipping, and self-speculative decoding, yielding up to 1.5$\times$ inference speedup with negligible accuracy loss. 
    Across more than 2400 training experiments, spanning models from 271M to 8.2B parameters and datasets up to 160B tokens, we demonstrate that these findings extend reliably to large-scale training regimes.
    All pre-training experiments were run on Cerebras CS-3 systems.
\end{abstract}

\begin{figure}[H]
    \centering
    \includegraphics[width=\columnwidth]{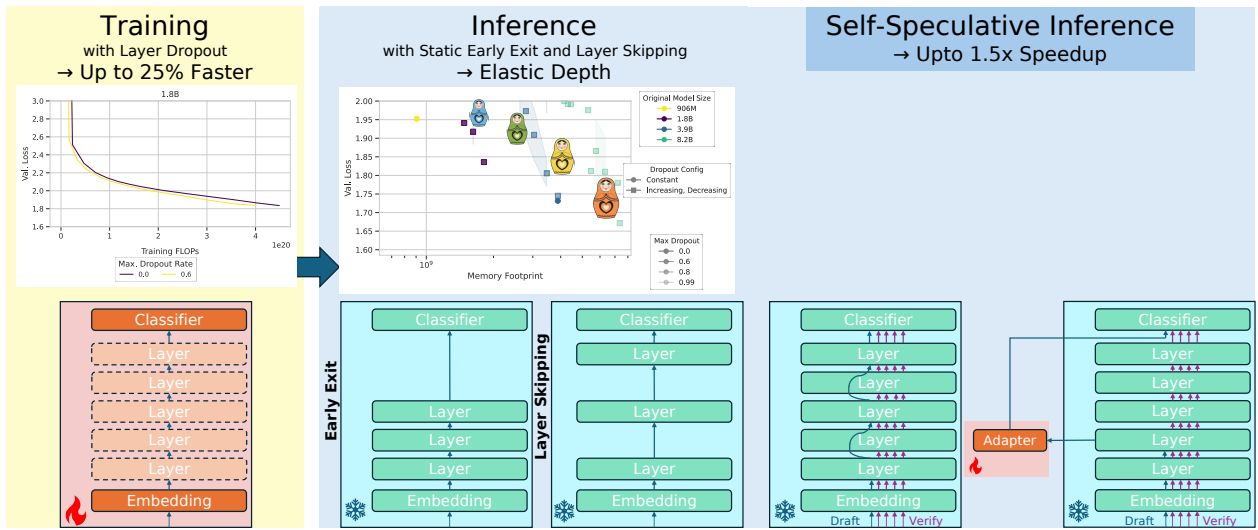}
    \caption{\textbf{Layer dropout as a unified mechanism for efficient LLM training and inference.} (\textit{Left}) Layer dropout skips layers stochastically during pre-training, leading to faster training, and with our proposed configuration, does so without sacrificing validation loss. (\textit{Center}) Trained models gain zero-shot ``elastic depth,'' degrading gracefully under early exit and layer skipping. (\textit{Right}) This robustness carries over to post-training adapters and self-speculative decoding for lossless inference speedup.}
    \label{fig:main}
\end{figure}

\section{Introduction}\label{sec:intro}

Pretraining large language models (LLMs) demands extraordinary computational resources~\citep{narayanan2021efficient,meng2025astral}, where small improvements in time-to-accuracy can save millions of dollars~\citep{coleman2019analysis,shen2024efficient} and reduce carbon emissions~\citep{acun2022holistic,wang2025catransformerscarbonawaretransformers}. 

Historically, regularization techniques improved validation accuracy for given training budgets by reducing overfitting and stabilizing optimization~\citep{moradi2020survey,wang2013fast,murugan2017regularization}. \emph{Dropout} was widely adopted in convolutional networks~\citep{hinton2012improvingneuralnetworkspreventing} and early transformers~\citep{vaswani2023attentionneed}.
However, as LLMs scaled to billions of parameters and trillions of tokens, dropout has been largely abandoned~\citep{Raschka_2025}. Models trained for a single epoch over massive datasets have little opportunity for classical overfitting, and empirical evidence suggests activation dropout degrades performance under these conditions~\citep{liu-etal-2025-drop-dropout}.

One type of dropout, \Gls{layer_dropout}, also known as \Gls{stochastic_depth}~\citep{stochastic_depth}, can provide benefits beyond regularization.
Unlike activation dropout or unstructured sparsity, which typically do not translate into wall-clock speedups due to sparse-kernel overheads, skipping entire transformer blocks yields structured sparsity that can reduce active training FLOPs almost linearly with the dropout rate~\citep{NEURIPS2020_a1140a3d_progressive_layer_dropout,filter_prune_or_layer_prune}.
It also encourages robustness to reduced-depth execution at inference time, enabling a single pretrained model to dynamically adapt to different latency and compute budgets without retraining. This supports zero-shot depth-wise optimizations such as elastic depth~\citep{Fan2020Reducing}, early exit~\citep{elhoushi-etal-2024-layerskip}, and intermediate layer skipping~\citep{stochastic_depth,Cai_2021_WACV_dynamic_routing}.

Despite layer dropout's promise, its role in state-of-the-art LLM pretraining has never been established through a comprehensive evaluation at scale. Existing evidence is fragmented across model families, dataset sizes, and implementation conventions. Many reported degradations may reflect suboptimal schedules or hyperparameters, rather than fundamental limitations. This leaves a basic unresolved question: \emph{should layer dropout be used in modern large-scale LLM training, and if so, how should it be configured to preserve accuracy while delivering training and deployment benefits?}

We provide the first unified experimental study of layer dropout in LLMs, systematically varying (i) optimizer hyperparameters, (ii) depth-wise distribution and granularity of layer sparsity, and (iii) temporal dropout schedules, across fixed architecture and data. Across 2400+ training runs spanning 271M to 3.9B parameters and up to 116B tokens, we identify configurations that reliably improve training and inference efficiency. Our contributions are:

\begin{enumerate}[nosep]
    \item \textbf{Improved Compute--Accuracy Trade-offs:} Properly configured layer dropout reduces training FLOPs while achieving validation loss competitive with, and in several cases superior to, dense baselines at scale.
    \item \textbf{Joint Optimization Framework:} We identify key interactions between dropout configurations, schedules, and optimizer hyperparameters that mitigate degradations observed in prior work.
    \item \textbf{Depth-Elastic Inference:} The \emph{average training dropout rate} predicts zero-shot robustness to early exit and layer skipping without retraining.
    \item \textbf{Scaling Analysis and Best Practices:} We analyze performance across model and data scales, recommending a progressively increasing distribution across \emph{depth} paired with a decreasing schedule across \emph{steps}, yielding up to 25\% training FLOPs savings and up to 1.5$\times$ inference speedup.
\end{enumerate}

\section{Related Work}\label{sec:related_work}

\paragraph{Dropout granularity and scope.}
Dropout encompasses a family of techniques that differ in the granularity at which stochastic sparsity is applied. Prior work distinguishes activation-level dropout, weight-level dropout (e.g., DropConnect~\citep{pmlr-v28-wan13-dropconnect}), and structured dropout that operates on groups of parameters such as channels, layers, or blocks~\citep{electronics12143106_dropout_regularization}. In this paper, we focus exclusively on \emph{structured, depth-wise dropout}—i.e., stochastic removal of entire transformer blocks during training—commonly referred to as \emph{layer dropout} or \emph{stochastic depth}~\citep{stochastic_depth}. We do not study neuron-level or weight-level dropout, which induce fine-grained sparsity and are known to interact differently with hardware efficiency and optimization dynamics.

\paragraph{Dropout in large-scale LLM pretraining.}
As language models scaled to billions of parameters and trillion-token datasets, explicit regularization techniques—including activation dropout—have largely disappeared from state-of-the-art pretraining recipes. Early decoder-only models such as GPT-3~\citep{gpt3} and OPT~\citep{opt} retained the dropout settings inherited from \citet{vaswani2023attentionneed}, while later models such as PaLM~\citep{palm} applied dropout only during finetuning, and LLaMA-style models no longer explicitly document its use. Recent empirical studies further suggest that activation dropout can degrade performance in single-epoch, large-data regimes~\citep{liu-etal-2025-drop-dropout}, reinforcing the prevailing view that dropout is unnecessary or harmful at scale. At the same time, dropout has been shown to remain beneficial in multi-epoch or data-limited settings~\citep{xue2023repeatrepeatinsightsscaling}, indicating that its utility is highly regime-dependent.

Importantly, techniques originally introduced as regularizers may persist in modern LLM training for reasons unrelated to overfitting prevention. Weight decay, for example, has been shown to primarily influence optimization dynamics rather than classical generalization in large-scale pretraining~\citep{d'angelo2024why_need_weight_decay}. This motivates re-examining dropout—particularly structured variants—without assuming that its value must stem from regularization in the traditional sense.

\paragraph{Layer dropout across scale.}
Layer dropout was originally proposed to stabilize optimization in very deep residual networks~\citep{stochastic_depth} and later become standard in large-scale vision models. However, its optimal strength has been observed to diminish as dataset scale increases: for example, ConvNeXt models trained on ImageNet-22K require substantially lower dropout rates than those trained on ImageNet-1K~\citep{convnext}. A similar pattern appears in language modeling. Progressive Layer Dropout~\citep{NEURIPS2020_a1140a3d_progressive_layer_dropout} and LayerDrop~\citep{Fan2020Reducing} reported improved convergence and robustness in BERT-era, multi-epoch settings on relatively small corpora. In contrast, more recent work applying layer dropout to decoder-only LLMs trained on large token budgets has reported non-negligible accuracy degradation~\citep{elhoushi-etal-2024-layerskip}, suggesting that naive extensions of earlier recipes may not transfer to modern regimes. To date, the literature lacks a controlled, large-scale evaluation that reconciles these conflicting findings by systematically varying dropout configurations and optimizer settings.

\paragraph{Training-aware approaches to depth elasticity.}
Layer dropout is closely related to a broader class of training-aware methods designed to enable inference-time efficiency. In compression, approaches such as Quantization-Aware Training (QAT) consistently outperform post-training quantization by exposing the model to reduced precision during optimization~\citep{stock2021qat_noise_training}. Analogously, depth-aware training aims to make models robust to reduced depth at inference time. Prior work has explored achieving depth elasticity via auxiliary losses, routers, or adapters added during or after pretraining, including early-exit models~\citep{balcony}, routing-based skipping~\citep{d-llm,MoD}, and hybrid speculative decoding schemes~\citep{draft-and-verify}. Other approaches train elastic architectures explicitly, such as Once-for-All~\citep{OnceForAll}, MatFormer~\citep{MatFormer}, and Nemotron-Elastic~\citep{NemotronElastic}. While effective, these methods typically introduce architectural changes, additional parameters, or auxiliary objectives.

Layer dropout occupies a distinct position within this landscape: it induces robustness to depth-wise inference optimizations directly during pretraining, without modifying the model architecture or introducing additional losses. Prior work demonstrated that this can enable elastic inference at small scales~\citep{Fan2020Reducing}, but whether similar benefits can be realized at modern LLM scales without sacrificing base-model accuracy has remained unresolved.

\section{Methodology}
In our experiments, we train decoder-only transformers following the architecture of Celerity models \citep{bergsma2025scalingcollapseefficientpredictable}: ALiBi position embeddings \citep{ALiBi}, squared ReLU activations \citep{relu2}, and Llama3 vocabulary \citep{llama3}. Specific architectural dimensions for all model sizes are detailed in the Appendix. Our datasets are obtained from a diverse corpus of natural language text and code.

To develop best practices and quantify the effects of layer dropout, we first identify optimal hyperparameters for each dropout rate (Sec.~\ref{sec:hp}). We then determine optimal granularity (Sec.~\ref{sec:dropout_granularity}) and configuration (Sec.~\ref{sec:dropout_configs}). Following \citet{chinchilla}, these experiments utilize a compute-optimal budget of 20 tokens-per-parameter (TPP) at each model size. Subsequently, we evaluate benefits across various depth-wise inference optimizations (Sec.~\ref{sec:inference_optimizations}), then quantify accuracy as training scales to larger datasets (Sec.~\ref{sec:training_scaling}). We conclude with larger-scale runs with aggressive dropout rate to demonstrate its final performance and inference advantages.

\section{Preliminary}
\paragraph{General Formulation of Layer Dropout} We start by denoting residual layer $\ell \in \{0, \dots, L-1\}$, of an $L$ layer neural network at training step $t \in \{0, \dots, T-1\}$, as:
\begin{equation}
    \HB^{\ell+1,t}   = \HB^{\ell,t} + f^{\ell}(\HB^{\ell,t})
\end{equation}
where, in the domain of natural language processing, activation tensor $\HB \in \mathbb{R}^{B \times S \times d}$, $B$ is batch size, $S$ is sequence length, $d$ is hidden dimension. 


When layer dropout is applied with rate $p^{l,t}$, the operation of the layer during training at step $t$ becomes:
\begin{equation}\label{eq:layer_dropout_train}
    \HB^{\ell+1,t}   = \HB^{\ell,t} + r^{\ell,t}_{\text{train}}\MB^{\ell,t} f^l(\HB^{\ell,t}) \\
\end{equation}
where mask $\MB^{\ell,t} \in \{0,1\}^{B} \sim \text{Bernoulli}(1-p^{\ell,t})$ is a Bernoulli random vector, and $r_{\text{train}}$ is a scaling factor applied during training. $r_{\text{train}}$ is defined differently in different layer dropout literature, and we will discuss our choice later.

The $b^\text{th}$ sequence of $\HB$ during training is now equal to\footnote{For neuron dropout, i.e., the default variant of dropout introduced by ~\citep{hinton2012improvingneuralnetworkspreventing}, $\MB \in \{0,1\}^{\{B \times S \times d\}}$.}:
\begin{equation}
\HB^{\ell+1,t}[b,:,:] = 
\begin{cases} 
    \HB^{\ell,t}[b,:,:], \\ 
    \hfill \text{with probability } p, \\[2mm]
    \HB^{\ell,t}[b,:,:] + r^{\ell,t}_{\text{train}} f^l\big(\HB^{\ell,t}[b,:,:]\big), \\ 
    \hfill \text{with probability } 1-p.
\end{cases}
\end{equation}
 While layer dropout could be implemented during training by executing $f(\HB^{\ell,t})$ on all sequences $b \in \{0, 1, ..., B-1\}$ of $\HB$, and multiplying its output by $\MB^{\ell,t}$, a more efficient implementation would be to only execute $f(\HB^{\ell,t})$ on sequences $b \in \{\, b_i \mid \MB^{\ell,t}[b_i] = 1 \,\}$. This leads to a saving a portion $p$ of training FLOPs of the layer. 

During inference, dropout is typically disabled and a distinct scaling factor, $r^{\ell}_{\text{eval}}$, is applied:
\begin{equation}\label{eq:layer_dropout_eval}
    \HB^{\ell+1}   = \HB^{\ell} + r^{\ell}_{\text{eval}}f^\ell(\HB^{\ell}) \\
\end{equation}

\paragraph{Layer Dropout for a Transformer} We denote the operation of layer $\ell \in \{0, \dots, L-1\}$ of an $L$, layer transformer model, at time step $t \in \{0, \dots, T-1\}$, during training as:
\begin{equation}
\begin{aligned}
    \ZB^{\ell,t}   &= \XB^{\ell,t} + f^l_{\text{attn}}(\XB^{\ell,t}) \\
    \XB^{\ell+1,t} &= \ZB^{\ell,t} + f^l_{\text{ffn}}(\ZB^{\ell,t})
\end{aligned}
\end{equation}
where $\XB, \ZB \in \mathbb{R}^{B \times S \times d}$,  $f^{\ell}_{\text{attn}}$ is the attention layer and $f^{\ell}_{\text{ffn}}$ is the feed-forward network (\Gls{FFN}).\footnote{This is a simplified form that does not refer to layer normalization or different variants of attention and FFN, but the subsequent formulation generalizes to different transformer variants that include pre-, post-, layer normalization, different variants or alternatives to attention, FFNs, and mixture of experts, as long as residual connection exists.} 

When layer dropout is applied with rate $p^{\ell,t}$, the operation at transformer layer, $\ell$, step, $t$, during training becomes:
\begin{equation}\label{eq:transformer_layer_dropout_train}
\begin{aligned}
    \ZB^{\ell,t}   &= \XB^{\ell,t} + r^{\ell,t}_{\text{train}}\MB^{\ell,t}_{\text{attn}} f^{\ell}_{\text{attn}}(\XB^{\ell,t}) \\
    \XB^{\ell+1,t} &= \ZB^{\ell,t} + r^{\ell,t}_{\text{train}}\MB^{\ell,t}_{\text{ffn}} f^{\ell}_{\text{ffn}}(\ZB^{\ell,t})
\end{aligned}
\end{equation}
and during inference becomes:
\begin{equation}\label{eq:transformer_layer_dropout_eval}
\begin{aligned}
    \ZB^{\ell,t}   &= \XB^{\ell,t} + r^{\ell,t}_{\text{eval}} f^{\ell}_{\text{attn}}(\XB^{\ell,t}) \\
    \XB^{\ell+1,t} &= \ZB^{\ell,t} + r^{\ell,t}_{\text{eval}} f^{\ell}_{\text{ffn}}(\ZB^{\ell,t})
\end{aligned}
\end{equation}

\section{Hyperparameters}\label{sec:hp}
\paragraph{Background} To avoid the ``hyperparameter lottery'' phenomenon \citep{cerebras2024mupguide}, and to ensure we compare against a strong baseline, we systematically optimize learning rate, batch size, and weight decay for each dropout rate before evaluating configurations. Prior literature offers varying strategies—from coupling dropout with $max$-norm regularization \citep{hinton2012improvingneuralnetworkspreventing} to using learning rates $10\times$ larger than baselines \citep{NEURIPS2020_a1140a3d_progressive_layer_dropout}—yet systematic consensus remains elusive. To our knowledge, this is the first study to perform joint optimization of these hyperparameters for layer-wise dropout. We tune a small model with dimensions depth $L_{\text{base}}$, width $d_{\text{base}}$ on dataset $D_{\text{base}}$ to determine learning rate $\eta_{\text{base}}$, weight decay $\lambda_{\text{base}}$, initialization $\sigma_{\text{base}}$, ans batch size $B_{\text{base}}$, then scale via $\mu$P \citep{mup}, CompleteP \citep{CompleteP}, and Power Lines \citep{PowerLines}.

\paragraph{Dropout Scale}
The scaling parameters $r_{\text{train}}$ and $r_{\text{eval}}$ from Equations~\ref{eq:layer_dropout_train} and ~\ref{eq:layer_dropout_eval} require careful consideration. We define layer density $\rho = 1 - p$. The choice of scaling parameters varies across different research work and frameworks. Moreover, they are occasionally left implicit in published papers, and we often need to inspect their source code to specify which scaling they use.
The original dropout paper \citep{hinton2012improvingneuralnetworkspreventing} used $r_{\text{train}}=1$, $r_{\text{eval}}=\rho$. Standard libraries like PyTorch and TensorFlow use $r_{\text{train}}=1/\rho$ for dropout. For layer dropout, the first stochastic depth paper \citep{stochastic_depth} used $r_{\text{train}}=1$, $r_{\text{eval}}=p$; DINOv2~\citep{dinov2}\footnotehyper{\label{footnote:dinov2}\url{https://github.com/facebookresearch/dinov2/blob/main/dinov2/layers/drop_path.py}} used $r_{\text{train}}=1/\rho$, $r_{\text{eval}}=1$; \texttt{fairseq}\footnotehyper{\label{footnote:fairseq}\url{https://github.com/facebookresearch/fairseq/blob/main/fairseq/modules/layer_drop.py}} (that implemented \citet{Fan2020Reducing}) and \texttt{torchtune}\footnotehyper{\label{footnote:torchtune}\url{https://github.com/meta-pytorch/torchtune/blob/main/torchtune/modules/layer_dropout.py}} (that implemented \citet{elhoushi-etal-2024-layerskip}) set both to 1. We demonstrate that selecting $r_{\text{train}}=1/\rho$ is critical for stable hyperparameter transfer.

\begin{figure}
  \centering
  \begin{subfigure}[t]{0.25\columnwidth}
    \centering
    \includegraphics[width=\linewidth]{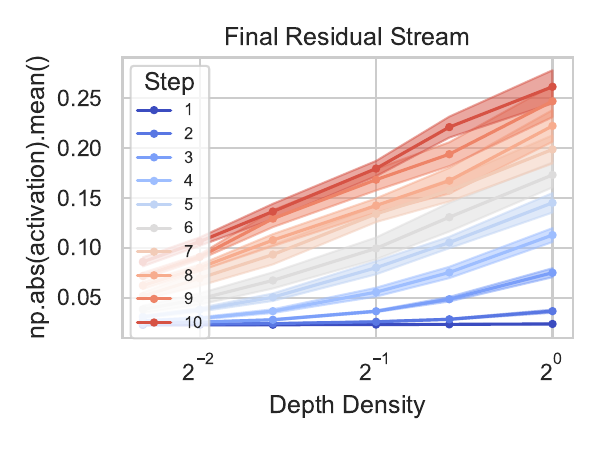}
    \caption{$r^{l,t}_{\text{train}}=1$}
    \label{fig:coordinate_checks_uniform_dropout:one}
  \end{subfigure}
  \begin{subfigure}[t]{0.25\columnwidth}
    \centering
    \includegraphics[width=\linewidth]{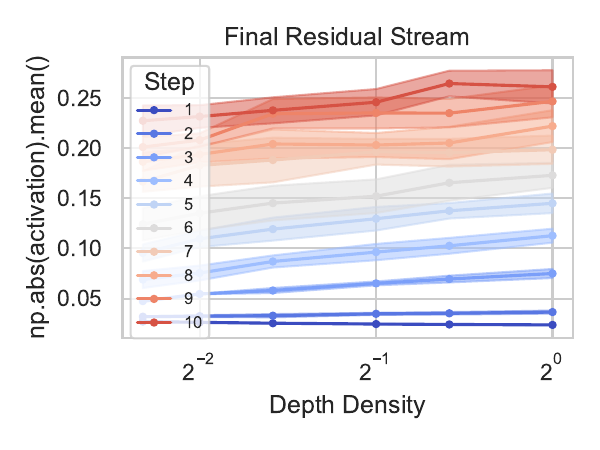}
    \caption{$r^{l,t}_{\text{train}}=1/\rho^{l,t}$ \crown}
    \label{fig:coordinate_checks_uniform_dropout:one_over_rho}
  \end{subfigure}

    \caption{Coordinate Check. Scaling with $r^{l,t}_{\text{train}}=1/\rho^{l,t}$ during training with layer dropout yields stable activation scale across depth density. More details in Sec.~\ref{sec:coordinate_check}.} 
    \label{fig:coordinate_checks_uniform_dropout}
    \vspace{-20pt}
\end{figure}

To determine the optimal scale factor $r_{\text{train}}$, we follow CompleteP's Maximal Residual Stream Update Desideratum \citep{CompleteP}, which facilitates hyperparameter transfer across model depths $L$.
\begin{des}
[Maximal Residual Stream Update]
\label{des:mu}
Each residual block's weights should contribute order $1/L$ to feature movements, and each non-residual block should contribute constant order. 
More precisely, for all $\ell \in [L-1]$, each block's parameter update $\bm\theta^{\ell} \mapsto \bm\theta^{\ell} + \Delta \bm\theta^{\ell}$ should contribute the change $\frac{1}{d}\| \Delta_{\bm \theta^{\ell}} \HB^{\ell+1} \|^2_2 \in \Theta( 1/L )$. 
Moreover, for the embedding and unembedding layers we require $\frac{1}{d} \| \Delta \mathbf W^0 \XB \|^2_2  \in \Theta(1)$ and $\frac{1}{d} \| \Delta \mathbf W^L \HB^L \|^2_2 \in \Theta(1)$. 
\end{des}
Since layer dropout reduces the effective depth of the network during training, we treat models with different dropout rates $\rho$ as having different effective depths, and apply this desideratum to ensure stable initialization across these effective depths. Our coordinate checks in Fig.~\ref{fig:coordinate_checks_uniform_dropout} empirically evaluate which scaling factor better satisfies stable initialization across dropout rates: $r_{\text{train}}=1$ fails, necessitating per-rate tuning, whereas $r_{\text{train}}=1/\rho$ largely satisfies these checks, enabling optimal hyperparameters transfer across many layer dropout rates.

\paragraph{Transfer Test}

\begin{figure*}
    \centering
    \begin{subfigure}[b]{0.32\textwidth}
        \centering
        \includegraphics[width=\linewidth]{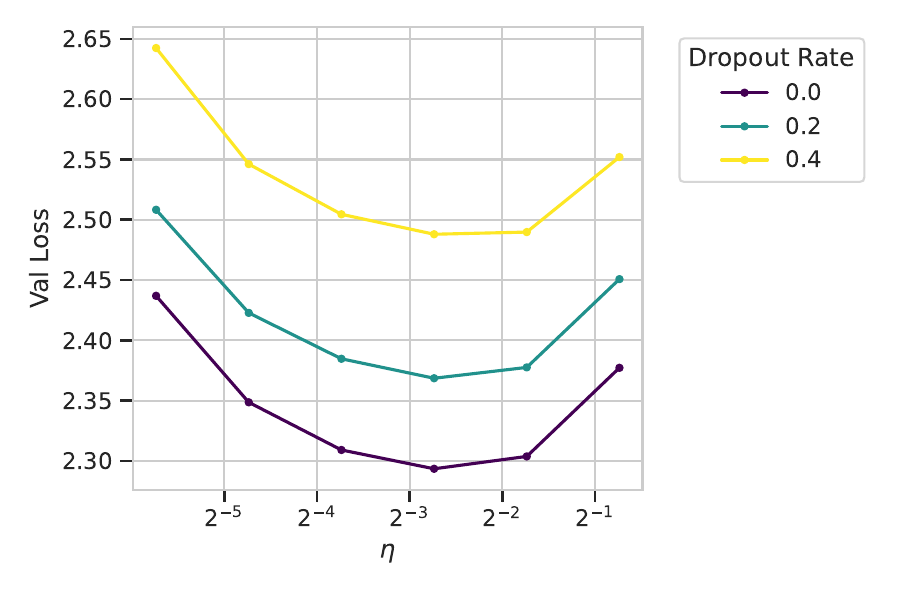}
        \caption{Learning rate.}
        \label{fig:20250821_depth_mup_transfer_lr}
    \end{subfigure}
    \hfill
    \begin{subfigure}[b]{0.32\textwidth}
        \centering
        \includegraphics[width=\linewidth]{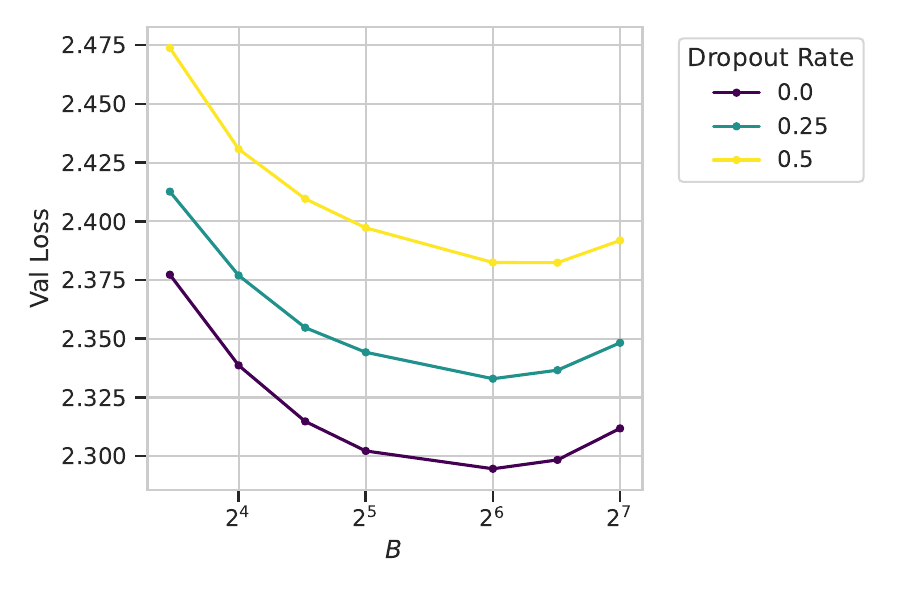}
        \caption{Batch size.}
        \label{fig:20250916_depth_mup_transfer_bs_zero_wd}
    \end{subfigure}
    \hfill
    \begin{subfigure}[b]{0.32\textwidth}
        \centering
        \includegraphics[width=\linewidth]{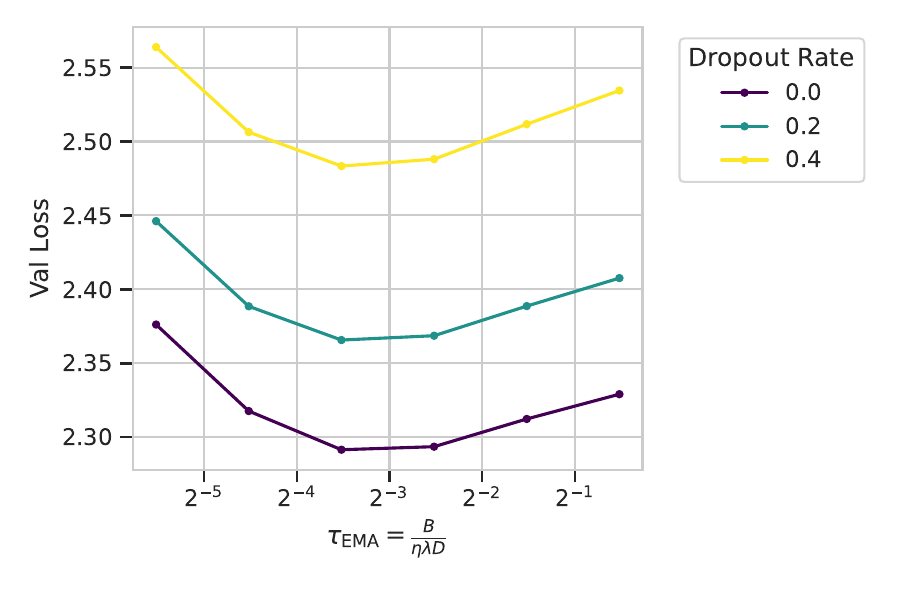}
        \caption{AdamW time scale.}
    \label{fig:20250924_depth_mup_transfer_tema_scale_tpp_bs128}
    \end{subfigure}
    \caption{Analysis of hyperparameter transferability on 271M model. We observe that optimal value for each hyperparameter remains similar across most layer dropout rates with the scaling factor $r_{\text{train}}=1/\rho$.}
    \label{fig:hyperparameter_transfer_test}
\end{figure*}

Fig.~\ref{fig:hyperparameter_transfer_test} verifies that $r_{\text{train}}=1/\rho$ enables hyperparameter transfer: optimal $\eta$, $\lambda$, and $B$ remain constant across dropout rates. Hence, we adopt Table~\ref{tab:parameterization-summary}'s transfer rules with $r_{\text{train}}=1/\rho$ for all our upcoming experiments. We set $r_{\text{eval}}=1$ to ensure that $\mathbf{H}^{\ell+1}_{\text{eval}} = \mathbb{E}[\mathbf{H}^{\ell+1}_{\text{train}}]$. 

\section{Dropout Granularity}\label{sec:dropout_granularity}
\subsection{Model Granularity}
\paragraph{Background} When layer dropout was first introduced by~\citep{stochastic_depth}, it was applied on residual blocks in CNNs, where each residual block consisted of two convolution-batchnorm pairs, separated by ReLU. In transformers, each layer consists of 2 residual blocks: an attention residual block followed by a FFN residual block. An open question is whether to apply layer dropout separately to attention and FFN (i.e., the Bernoulli mask tensors $\MB^{\ell,t}_{\text{attn}}$ and $\MB^{\ell,t}_{\text{ffn}}$ are sampled independently at each training step, $t$), which we refer to as \Gls{sub_layer_dropout}, or to apply it on the whole transformer layer (i.e., $\MB^{\ell,t}_{\text{attn}} = \MB^{\ell,t}_{\text{ffn}} \quad \forall \ell$), which we refer to as \Gls{layer_dropout}. Different research work have used different types: DINOv2~\citep{dinov2} and \citep{NEURIPS2020_a1140a3d_progressive_layer_dropout} used sub-layer dropout, while LayerDrop~\citep{Fan2020Reducing} and LayerSkip~\citep{elhoushi-etal-2024-layerskip} used layer dropout. However, to the best of our knowledge, we are the first to systematically evaluate a comparison between them.

\paragraph{Analysis} In Table~\ref{tab:model_granularity} we compare layer dropout and sub-layer dropout at various model sizes. The results clearly show that in terms of accuracy, Layer Dropout is better. Note that as model size increases, loss degradation introduced by dropout diminishes, which will later encourage us to try larger dropout rates for larger models. This may be contrary to the notion that finer grain sparsity leads to higher accuracy, but could be explained by other research work that show that attention and FFN work in tandem~\citep{agarwal2026gradientdynamicsattentioncrossentropy}. We leave investigating the reason sub-layer dropout underperforms layer dropout, and also leave investigating other configurations such as applying dropout only on attention or only on FFN, for future work.

\begin{table}[h]
\centering
\small
\caption{Ablating model granularities. Models trained at 20 TPP.}
\label{tab:model_granularity}
\begin{tabular}{@{}llccccc@{}}
\toprule
\makecell[l]{\textbf{Model}\\\textbf{Size}} & \makecell[l]{\textbf{Dropout}\\\textbf{Type}} & \makecell[l]{\textbf{Dropout}\\\textbf{Rate}} & \makecell{\textbf{Train}\\\textbf{Loss} $\downarrow$} & \textbf{\% $\Delta$ $\downarrow$} & \makecell{\textbf{Val}\\\textbf{Loss} $\downarrow$} & \textbf{\% $\Delta$ $\downarrow$} \\ \midrule
271M & Baseline & --- & 2.293 & 0.00\% & 2.294 & 0.00\% \\
     & SubLayer & 0.1 & 2.421 & 4.72\% & 2.377 & 3.61\% \\
     & Layer \crown & 0.1 & \textbf{2.419} & \textbf{3.95\%} & \textbf{2.367} & \textbf{3.18\%} \\ \midrule
503M & Baseline & --- & 2.140 & 0.00\% & 2.110 & 0.00\% \\
     & SubLayer & 0.1 & 2.260 & 3.94\% & 2.174 & 3.03\% \\
     & Layer \crown & 0.1 & \textbf{2.178} & \textbf{2.60\%} & \textbf{2.170} & \textbf{2.84\%} \\ \bottomrule
\end{tabular}
\end{table}

\finding{Layer dropout that drops whole transformer blocks for each sample, leads to higher accuracy results than sub-layer dropout that drops attention and FFN sub-blocks separately.}

\subsection{Tensor Granularity}
\paragraph{Background} The next open question we tackle is whether it is better to apply layer dropout at batch granularity, i.e. $\MB^{\ell,t}[b] = M^{\ell,t}$ $\forall b$, where $M^{\ell,t}\sim\text{Bernoulli}(1-p^{\ell,t})$ is drawn once per layer $\ell$ and step $t$ (so $\MB^{\ell,t}[b]$ takes the same value for all sequences $b$), or at sequence granularity, i.e. $\MB^{\ell,t}[b] \sim \text{Bernoulli}(1-p^{\ell,t})$ drawn i.i.d. for each $b$ (so $\MB^{\ell,t}[b]$ is sampled independently for each sequence $b$).

In literature, this does not seem to have been discussed, and we usually need to resort to the codebases of different papers to find out which type each has used. The implementation of the pioneer Stochastic Depth paper\footnotehyper{\label{footnote:stochastic_depth}\url{https://github.com/yueatsprograms/Stochastic_Depth/blob/master/ResidualDrop.lua}} as well as the \texttt{fairseq}\footnotehyper{\label{footnote:fairseq}\url{https://github.com/facebookresearch/fairseq/blob/main/fairseq/modules/layer_drop.py}} implementation of LayerDrop used per-batch layer dropout, while DINOv2~\citep{dinov2}\footnotehyper{\label{footnote:dinov2}\url{https://github.com/facebookresearch/dinov2/blob/main/dinov2/layers/drop_path.py}}, \texttt{timm}\footnotehyper{\label{footnote:timm}\url{https://github.com/huggingface/pytorch-image-models/blob/main/timm/layers/drop.py}}, and \texttt{torchtune}\footnotehyper{\label{footnote:torchtune}\url{https://github.com/meta-pytorch/torchtune/blob/main/torchtune/modules/layer_dropout.py}} implementation of LayerSkip used per-sequence. To the best of our knowledge, we are the first to systematically compare per-batch and per-sequence layer dropout.

\begin{figure}
    \centering
    \includegraphics[width=0.5\linewidth]{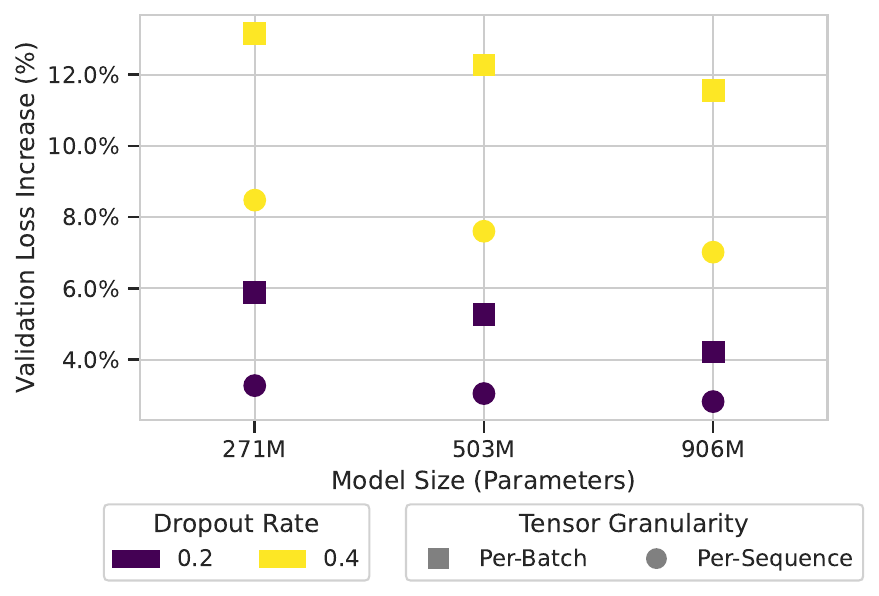}
    \caption{Ablating tensor granularity. Models trained at 20 TPP.}
    \label{fig:20250626_drop_layer_per_row__dropout_activation_granularity}
\end{figure}

\paragraph{Analysis} Fig.~\ref{fig:20250626_drop_layer_per_row__dropout_activation_granularity} compares the accuracy results of applying layer dropout per batch and per sequence. The results clearly show that per-sequence leads to better losses. This is in line with the notion that finer grain sparsity leads to higher accuracy. In terms of compute performance, per-batch layer dropout has the advantage of not having to load the weights of a layer during a training step. However, if training is compute bound (i.e., batch size and context length are large enough), per-sequence dropout should lead to speedup similar to per-batch dropout as both save the same compute FLOPs. A middle ground that could combine the benefits of not loading weights of per-batch dropout and fine-grain sparsity of per-sequence dropout, could be satisfied in distributed training where each device drops different batches, or training with gradient accumulation where a different mini-batch is dropped per gradient accumulation step. We leave exploring such approaches, as well as comparing with even finer-grain dropout such as per-token or per-neuron, for future work.

\finding{For any given layer dropout rate, dropout per-sequence leads to lower loss than dropout per-batch.}

\section{Dropout Configurations}\label{sec:dropout_configs}
\subsection{Dropout Distribution}

\paragraph{Background} Various dropout distributions across layers have been proposed to optimize training efficiency and model depth. We formalize three primary distributions for dropout rate $p$ at layer $\ell$:

\begin{enumerate}[nosep]
    \item \textbf{\Gls{uniform_distribution}}: where all layers have the same dropout rate, $p^{\ell,t}_{\text{uniform}} = p_{\text{max}}$.
    \item \textbf{\Gls{increasing_layer_distribution}}: where dropout rate starts at 0 at the first layer and linearly increases across layers to reach $p_{\text{max}}$ at the last layer, $p^{\ell,t}_{\text{ILD}} = \frac{\ell}{L-1} \cdot p_{\text{max}}$ \citep{stochastic_depth, dinov2, NEURIPS2020_a1140a3d_progressive_layer_dropout, elhoushi-etal-2024-layerskip}.
    \item \textbf{\Gls{alternating_layer_distribution}}: where layer dropout is only applied at every other layer, $p^{\ell,t}_{\text{ALD}} = p_{\text{max}} \cdot \mathbf{1}_{\ell\equiv 1 \text{(mod 2)}}$ \citep{Fan2020Reducing}.
\end{enumerate}
where $p_{\text{max}}$ is the maximum dropout rate. Note that $p^{\ell,t}_{\text{ALD}} \in \{0, p_{\text{max}} \}$, whereas $p^{\ell,t}_{\text{ILD}} \in [0, p_{\text{max}}]$.

For any distribution, the average dropout and corresponding nonembedding FLOPs\footnote{For the remaining of the paper, we use the term FLOPs to refer to nonembedding FLOPs.} savings at step $t$ are defined as:
\begin{equation}
    p^{t}_{\text{mean}} = \text{FLOPs Savings}^{t} = \frac{1}{L}\textstyle\sum_{\ell=0}^{L-1}p^{\ell,t}
\end{equation}
Mathematically, $p^{t}_{\text{ILD}_{\text{mean}}} = 0.5p_{\text{max}}$\footnote{Follows from applying the arithmetic series formula $\sum_{i=0}^{n-1} a_i = \frac{n}{2}(a_0 + a_{n-1})$ to the per-step mean $p^t_{\text{ILD}_{\text{mean}}} = \frac{1}{L}\sum_{\ell=0}^{L-1} p^{\ell,t}_{\text{ILD}} = \frac{1}{L}\sum_{\ell=0}^{L-1} \frac{\ell}{L-1}\cdot p_{\max} = \frac{p_{\max}}{L}\cdot\frac{L}{2}\left(\frac{0}{L-1} + \frac{L-1}{L-1}\right) = \frac{1}{2}\,p_{\max}$.} and $p^{t}_{\text{ALD}_{\text{mean}}} = \frac{\lfloor L/2 \rfloor}{L} p_{\text{max}}$, which is $\approx 0.5p_{\text{max}}$ for typical $L$. To the best of our knowledge, this study is the first to systematically analyze the differences between these layer dropout distributions at fixed FLOPs budgets.

\paragraph{Analysis} In Table~\ref{tab:dropout_distribution}, we compare uniform, ILD, and ALD grouped by equivalent FLOPs savings, finding that non-uniform distributions consistently outperform uniform ones under a fixed average dropout (as well as fixed FLOPs budget). While ALD is superior at the smallest model size, its advantage diminishes with scale, whereas ILD’s improvement over uniform widens. Although our ALD results with $p_{\text{max}}=0.2$ do not beat the baseline as reported in the multi-epoch regime of LayerDrop~\citep{Fan2020Reducing}, the observed reduction in dropout-induced degradation as models grow encourages further investigation at larger scales.
\def\baselineA{2.29358197804906}  
\def\baselineB{2.1095672428196}  
\def\baselineC{1.95261678741309} 

\newcommand{\lossfmtA}[1]{%
  \begingroup
    \edef\val{#1}%
    \ifnum \fpeval{(\val-\baselineA)<0}=1
      \textcolor{green!60!black}{%
        \num[
          round-mode=places,
          round-precision=3,
          minimum-decimal-digits=3
        ]{#1}%
      }%
    \else
      \num[
        round-mode=places,
        round-precision=3,
        minimum-decimal-digits=3
      ]{#1}%
    \fi
  \endgroup
}

\newcommand{\pctdeltaA}[1]{%
  \begingroup
    \sisetup{
      round-mode=places,
      round-precision=2,
      minimum-decimal-digits=2
    }%
    \edef\val{#1}%
    \edef\res{\fpeval{((\val-\baselineA)/\baselineA)*100}}%
    \ifnum\fpeval{(\val-\baselineA)<0}=1
      \textcolor{green!60!black}{\num{\res}\%}%
    \else
      \num{\res}\%%
    \fi
  \endgroup
}

\newcommand{\lossfmtB}[1]{%
  \begingroup
    \edef\val{#1}%
    \ifnum \fpeval{(\val-\baselineB)<0}=1
      \textcolor{green!60!black}{%
        \num[
          round-mode=places,
          round-precision=3,
          minimum-decimal-digits=3
        ]{#1}%
      }%
    \else
      \num[
        round-mode=places,
        round-precision=3,
        minimum-decimal-digits=3
      ]{#1}%
    \fi
  \endgroup
}

\newcommand{\pctdeltaB}[1]{%
  \begingroup
    \sisetup{
      round-mode=places,
      round-precision=2,
      minimum-decimal-digits=2
    }%
    \edef\val{#1}%
    \edef\res{\fpeval{((\val-\baselineB)/\baselineB)*100}}%
    \ifnum\fpeval{(\val-\baselineB)<0}=1
      \textcolor{green!60!black}{\num{\res}\%}%
    \else
      \num{\res}\%%
    \fi
  \endgroup
}

\newcommand{\lossfmtC}[1]{%
  \begingroup
    \edef\val{#1}%
    \ifnum \fpeval{(\val-\baselineC)<0}=1
      \textcolor{green!60!black}{%
        \num[
          round-mode=places,
          round-precision=3,
          minimum-decimal-digits=3
        ]{#1}%
      }%
    \else
      \num[
        round-mode=places,
        round-precision=3,
        minimum-decimal-digits=3
      ]{#1}%
    \fi
  \endgroup
}

\newcommand{\pctdeltaC}[1]{%
  \begingroup
    \sisetup{
      round-mode=places,
      round-precision=2,
      minimum-decimal-digits=2
    }%
    \edef\val{#1}%
    \edef\res{\fpeval{((\val-\baselineC)/\baselineC)*100}}%
    \ifnum\fpeval{(\val-\baselineC)<0}=1
      \textcolor{green!60!black}{\num{\res}\%}%
    \else
      \num{\res}\%%
    \fi
  \endgroup
}

\newcommand{\best}[1]{\textbf{#1}}

\begin{table}[h]
\small
\centering
\caption{Analysis of Dropout Distributions across layers. Models trained at 20 TPP.}
\label{tab:dropout_distribution}
\label{tab:dropout_distribution}
\begin{tabular}{@{}ccccll@{}}
\toprule
\makecell[t]{\textbf{Model}} &
\makecell[t]{\textbf{Training}\\\textbf{FLOPs}\\\textbf{Savings}} &
\makecell[t]{\textbf{Max Rate}} &
\makecell[t]{\textbf{Dropout}\\\textbf{Distb.}} &
\makecell[t]{\textbf{Val}} &
\makecell[t]{\textbf{\% $\Delta$}} \\ \midrule
\multirow{6}{*}{271M} 
    & 0\% 
    & - 
    & - 
    & \lossfmtA{2.29358197804906} 
    & \pctdeltaA{2.29358197804906} 
\\
\cmidrule(lr){2-6}
    & \multirow{3}{*}{10\%} 
    & 0.1 
    & Uniform 
    & \lossfmtA{2.33075255811499} 
    & \pctdeltaA{2.33075255811499} 
\\
    &  
    & 0.2 
    & Alternating \crown
    & \best{\lossfmtA{2.32309182427906}}
    & \best{\pctdeltaA{2.32309182427906}} 
\\
    &  
    & 0.2 
    & Increasing 
    & \lossfmtA{2.32842825049735}
    & \pctdeltaA{2.32842825049735} 
\\ 
\cmidrule(lr){2-6}
    & \multirow{3}{*}{20\%} 
    & 0.2 
    & Uniform 
    & \lossfmtA{2.368637159} 
    & \pctdeltaA{2.368637159} 
\\
    &  
    & 0.4 
    & Alternating \crown
    & \best{\lossfmtA{2.356840203}}
    & \best{\pctdeltaA{2.356840203}}
\\
    &
    & 0.4
    & Increasing
    & \lossfmtA{2.362629457}
    & \pctdeltaA{2.362629457} 
\\ 
\midrule
    \multirow{6}{*}{503M} 
    & 0\% 
    & - 
    & - 
    & \lossfmtB{2.1095672428196} 
    & \pctdeltaB{2.1095672428196} 
\\
\cmidrule(lr){2-6}
    & \multirow{3}{*}{10\%} 
    & 0.1 
    & Uniform 
    & \lossfmtB{2.140856052} 
    & \pctdeltaB{2.140856052} 
\\
    & 
    & 0.2 
    & Alternating \crown
    & \lossfmtB{2.142634623} 
    & \pctdeltaB{2.142634623} 
\\
    &
    & 0.2 
    & Increasing 
    & \best{\lossfmtB{2.132281742}}
    & \best{\pctdeltaB{2.132281742}}
\\ 
\cmidrule(lr){2-6}
    & \multirow{3}{*}{20\%} 
    & 0.2 
    & Uniform 
    & \lossfmtB{2.173887443} 
    & \pctdeltaB{2.173887443} 
\\
    &  
    & 0.4 
    & Alternating 
    & \lossfmtB{2.168975638} 
    & \pctdeltaB{2.168975638} 
\\    
    &  
    & 0.4 
    & Increasing \crown
    & \best{\lossfmtB{2.16195533}}
    & \best{\pctdeltaB{2.16195533}}
\\ 
\midrule
    \multirow{4}{*}{906M} 
    & 0\% 
    & - 
    & - 
    & \lossfmtC{1.95261678741309} 
    & \pctdeltaC{1.95261678741309} 
\\
\cmidrule(lr){2-6}
    & \multirow{2}{*}{10\%} 
    & 0.1 
    & Uniform 
    & \lossfmtC{1.977312397} 
    & \pctdeltaC{1.977312397} 
\\
    &
    & 0.2 
    & Alternating 
    & \lossfmtC{1.979108732} 
    & \pctdeltaC{1.979108732}
\\
    & 
    & 0.2 
    & Increasing \crown
    & \best{\lossfmtC{1.972352956}}
    & \best{\pctdeltaC{1.972352956}}
\\ 
\cmidrule(lr){2-6}
    & \multirow{2}{*}{20\%} 
    & 0.2 
    & Uniform 
    & \lossfmtC{2.007753786} 
    & \pctdeltaC{2.007753786}
\\ 
    &  
    & 0.4 
    & Alternating 
    & \lossfmtC{2.004482484} 
    & \pctdeltaC{2.004482484}
\\ 
    &  
    & 0.4 
    & Increasing \crown
    & \best{\lossfmtC{1.998349022}}
    & \best{\pctdeltaC{1.998349022}}
\\ 
\bottomrule
\end{tabular}
\end{table}



\finding{For the same training FLOPs budget, non-uniform dropout distribution across layers is better than uniform. As a model scales, ILD is recommended.}

\subsection{Dropout Schedule}\label{sec:dropout_schedule}

\paragraph{Background} While distributions govern sparsity across depth, the temporal schedule determines how regularization pressure evolves throughout pre-training. \citet{hillier2024stlmengineeringreportdropout} found decreasing schedules were better for LLM pre-training, whereas increasing schedules were superior for fine-tuning; however, \citet{liu-etal-2025-drop-dropout} recently claimed both fail in modern regimes. We formalize various time schedules for dropout rate $p$ at step $t$ over total duration $T$, where $p^{\ell}_{\text{dist}}$ represents a chosen layer distribution:

\begin{enumerate}[nosep]
    \item \textbf{\Gls{constant_time_schedule}}: where dropout rate is constant throughout training steps, $p^{\ell,t}_{\text{constant}} = p^{\ell}_{\text{dist}}$.
    \item \textbf{\Gls{increasing_time_schedule}}: where dropout rate starts at 0 at the beginning of training and linearly increases to $p^{\ell}_{\text{dist}}$ at the end of training, $p^{\ell,t}_{\text{ITS}} = p^{\ell}_{\text{dist}} \cdot \left( \frac{t}{T-1} \right)$.
    \item \textbf{\Gls{decreasing_time_schedule}}: where dropout rate starts at $p^{\ell}_{\text{dist}}$ at the beginning of training and linearly decreases to 0 at the end of training, $p^{\ell,t}_{\text{DTS}} = p^{\ell}_{\text{dist}} \cdot \left( 1 - \frac{t}{T-1} \right)$.
\end{enumerate}

To compare these schedules fairly, we define the mean training dropout $\bar{P}$ as the average rate across depth and time, representing total active training FLOPs savings:
\begin{equation}
    \bar{P} = \text{FLOPs Savings}_{\text{total}} = \frac{1}{T} \textstyle\sum_{t=0}^{T-1} \left( \frac{1}{L} \textstyle\sum_{\ell=0}^{L-1} p^{\ell,t} \right)
\end{equation}
Consequentially, $\bar{P}_{\text{uniform},\text{ITS}}=0.5p_{\text{max}}$ and $\bar{P}_{\text{ILD},\text{ITS}}=\bar{P}_{\text{ILD},\text{DTS}}=0.25p_{\text{max}}$\footnote{$\bar{P}_{\text{ILD,DTS}} = \frac{1}{LT}\sum_{t=0}^{T-1}\sum_{l=0}^{L-1}p^{l,t}_{\text{ILD,DTS}}$, where $p^{l,t}_{\text{ILD}} = p^t \frac{l}{L-1}$ and $p^{l,t}_{\text{DTS}} = p^l\left(1 - \frac{t}{T-1}\right)$. Separating the double sum into independent factors and applying the arithmetic series formula $\sum_{i=0}^{n-1} a_i = \frac{n}{2}(a_0 + a_{n-1})$ to each, the inner sum over $l$ evaluates to $\frac{L}{2}(0+1)\cdot p_{\max}$ and the outer sum over $t$ evaluates to $\frac{T}{2}(1+0)$, yielding the closed form $P_{\text{ILD,DTS}} = \frac{1}{4}\cdot p_{\max}$.}.

\paragraph{Analysis} In Table~\ref{tab:dropout_schedule}, we group configurations by total FLOPs savings. Across all scales, decreasing schedules consistently outperforms constant and increasing schedules. Notably, at 5\% FLOPs savings for 503M \& 906M, combined ILD and DTS achieves lower validation losses than the dense baseline, demonstrating for the first time that it is possible to beat the dense baseline with fewer training FLOPs.


\begin{table}[H]
\centering
\small
\setlength{\tabcolsep}{6pt}
\renewcommand{\arraystretch}{1.15}

\begin{tabular}{l c c c r r r r r r}
\toprule
\multirow[t]{2}{*}{\makecell[t]{\textbf{Training}\\\textbf{FLOPs}\\\textbf{Savings}}} &
\multirow[t]{2}{*}{\makecell[t]{\textbf{Max}\\\textbf{Dropout}}} &
\multirow[t]{2}{*}{\makecell[t]{\textbf{Layer}\\\textbf{Dist.}}} &
\multirow[t]{2}{*}{\makecell[t]{\textbf{Time}\\\textbf{Sched.}}} &
\multicolumn{2}{c}{\makecell[t]{\textbf{271M}}} &
\multicolumn{2}{c}{\makecell[t]{\textbf{503M}}} &
\multicolumn{2}{c}{\makecell[t]{\textbf{906M}}} \\
\cmidrule(lr){5-6}\cmidrule(lr){7-8}\cmidrule(lr){9-10}
& & & &
\makecell[t]{\textbf{Val}} & \makecell[t]{\textbf{\% $\Delta$}} &
\makecell[t]{\textbf{Val}} & \makecell[t]{\textbf{\% $\Delta$}} &
\makecell[t]{\textbf{Val}} & \makecell[t]{\textbf{\% $\Delta$}} \\
\midrule

\textbf{0\%} & --   & --     & --             & \lossfmtA{2.29358197804906} & \pctdeltaA{2.29358197804906} & \lossfmtB{2.1095672428196} & \pctdeltaB{2.1095672428196} & \lossfmtC{1.95261678741309} & \pctdeltaC{1.95261678741309} \\
\midrule

\textbf{5\%} & 0.05 & --     & --             & \lossfmtA{2.31123401933626} & \pctdeltaA{2.31123401933626} & \lossfmtB{2.12372084807234} & \pctdeltaB{2.12372084807234} & \lossfmtC{1.96314872109718} & \pctdeltaC{1.96314872109718} \\
             & 0.1  & Increasing & --             & \lossfmtA{2.31034808967545} & \pctdeltaA{2.31034808967545} & \lossfmtB{2.12130509495164} & \pctdeltaB{2.12130509495164} & \lossfmtC{1.9607470495025254} & \pctdeltaC{1.9607470495025254} \\
             & 0.2  & Increasing & Increasing         & \lossfmtA{2.32469393971243} & \pctdeltaA{2.32469393971243} & \lossfmtB{2.13875295709004} & \pctdeltaB{2.13875295709004} & \lossfmtC{1.97876957228436} & \pctdeltaC{1.97876957228436} \\
             & 0.2  & Increasing & Decreasing~\crown& \best{\lossfmtA{2.30409734810621}} & \best{\pctdeltaA{2.30409734810621}} & \best{\lossfmtB{2.11030398176934}} & \best{\pctdeltaB{2.11030398176934}} & \best{\lossfmtC{1.95134776117699}} & \best{\pctdeltaC{1.95134776117699}} \\
\midrule

\textbf{10\%} & 0.1  & --     & --             & \lossfmtA{2.33075255811499} & \pctdeltaA{2.33075255811499} & \lossfmtB{2.14085605225621} & \pctdeltaB{2.14085605225621} & \lossfmtC{1.97731239734844} & \pctdeltaC{1.97731239734844} \\
              & 0.2  & Increasing & --             & \lossfmtA{2.32842825049735} & \pctdeltaA{2.32842825049735} & \lossfmtB{2.13228174215425} & \pctdeltaB{2.13228174215425} & \lossfmtC{1.97235295621718} & \pctdeltaC{1.97235295621718} \\
              & 0.4  & Increasing & Increasing         & \lossfmtA{2.35674475184911} & \pctdeltaA{2.35674475184911} & \lossfmtB{2.17174235472712} & \pctdeltaB{2.17174235472712} & \lossfmtC{2.0140953658426906} & \pctdeltaC{2.0140953658426906} \\
              & 0.4  & Increasing & Decreasing~\crown& \best{\lossfmtA{2.31244438414495}} & \best{\pctdeltaA{2.31244438414495}} & \best{\lossfmtB{2.1155000555585}} & \best{\pctdeltaB{2.1155000555585}} & \best{\lossfmtC{1.95463081783972}} & \best{\pctdeltaC{1.95463081783972}} \\
\midrule

\textbf{20\%} & 0.2  & --     & --             & \lossfmtA{2.36863715917363} & \pctdeltaA{2.36863715917363} & \lossfmtB{2.17388744318757} & \pctdeltaB{2.17388744318757} & \lossfmtC{2.00775378597093} & \pctdeltaC{2.00775378597093} \\
              & 0.4  & Increasing & --             & \lossfmtA{2.36262945739719} & \pctdeltaA{2.36262945739719} & \lossfmtB{2.16195533004357} & \pctdeltaB{2.16195533004357} & \lossfmtC{1.99834902194001} & \pctdeltaC{1.99834902194001} \\
              & 0.8  & Increasing & Increasing         & \lossfmtA{2.44176668876159} & \pctdeltaA{2.44176668876159} & \lossfmtB{2.25708131141196} & \pctdeltaB{2.25708131141196} & \lossfmtC{2.09016815523465} & \pctdeltaC{2.09016815523465} \\
              & 0.8  & Increasing & Decreasing~\crown& \best{\lossfmtA{2.36086325814025}} & \best{\pctdeltaA{2.36086325814025}} & \best{\lossfmtB{2.14650904796403}} & \best{\pctdeltaB{2.14650904796403}} & \best{\lossfmtC{1.9829187944555207}} & \best{\pctdeltaC{1.9829187944555207}} \\
\bottomrule
\end{tabular}

\caption{Ablating dropout time schedule. We group different dropout configurations that have the same active non-embedding FLOPs reduction induced by layer dropout across all training steps. Models trained at 20 TPP.}
\label{tab:dropout_schedule}

\end{table}

Conversely, increasing schedule significantly degrades loss. While~\citep{NEURIPS2020_a1140a3d_progressive_layer_dropout} reported positive results with an exponential increasing schedule, we do not observe benefits in our large-scale single-epoch regime. We hypothesize the decreasing schedule's effectiveness stems from high initial noise at the beginning of training forcing weight space exploration (reducing bias), while subsequent decay allows settling into a stable minimum (reducing variance). This can also be viewed as a form of stochastic model growing, where effective capacity increases smoothly throughout training without explicit re-initialization of conventional model growing (e.g., \citep{ScalingSmart}). It can also be viewed as a form of curriculum learning~\citep{wang2021surveycurriculumlearning}, that starts training with a hard task of learning using small effective depth and the learning task gradually becomes easier as effective depth increases.

We leave other schedules such as applying dropout to mid-training, SFT, or continual pre-training for future work.

\finding{For a fixed training FLOPs budget, a schedule decaying from a maximum rate to zero consistently achieves the highest accuracy, outperforming both constant and increasing schedules across all scales.}

\takeaway{The best recommended practice for layer dropout configuration is increasing dropout across layers and decreasing dropout across time. This leads to the best accuracy for a given training FLOPs budget.}


\section{Inference Optimizations}\label{sec:inference_optimizations}
A primary motivation for pre-training with layer dropout is to induce robustness to depth-wise optimizations, including early exit, layer skipping, and layer pruning. We explore techniques that keep pre-trained weights intact. We categorize such techniques into ``Zero-Shot'' inference approaches that merely apply autoregressive decoding inference on a model with fewer layers without any modifications, and ``Post-Training'' approaches that add adapters or routers (albeit not modifying the model's weights) or modify the inference decoding algorithm.  We leave pruning approaches that require fine-tuning or weight modification, e.g., \citet{ShearedLlama, lu2024reassessing}, for future work.

\subsection{Zero-Shot Inference Benefits}


\subsubsection{Early Exit}
\begin{figure}[h]
    \centering
    \includegraphics[width=\columnwidth]{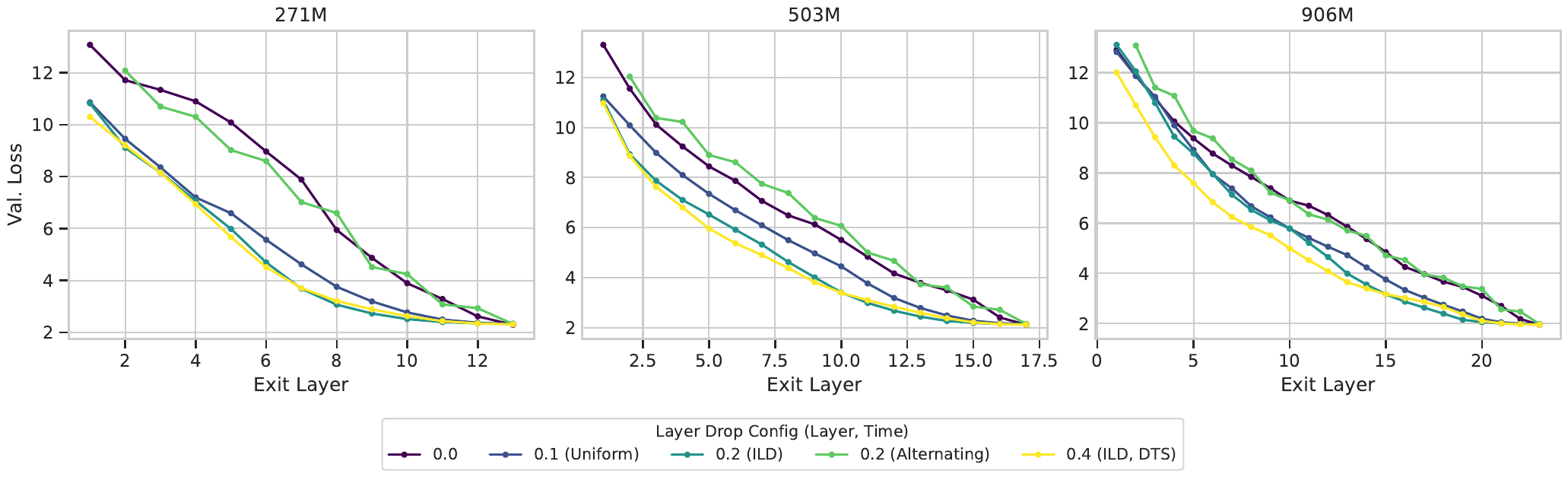}
    \caption{Early-exit validation loss for different model sizes trained at 20 TPP: no dropout vs. dropout configurations with 10\% FLOPs savings. Additional comparisons in Fig.~\ref{fig:early-exit-comparison-detailed}.}
    \label{fig:early-exit-comparison}
\end{figure}

We define early exit at layer $\ell'$ as executing the embedding layer, transformer layers $0$ to $\ell'-1$, and the unembedding layer. This ``static early exit'' is equivalent to skipping layers $\ell'$ to $L-1$. Fig.~\ref{fig:early-exit-comparison} and~\ref{fig:early-exit-comparison-detailed} illustrate results for various models. We see that for a model trained without dropout, loss deteriorates significantly even after exiting one layer earlier, but for a model trained with dropout, loss remains steady when exiting early for a portion of layers. Throughout all layers, exiting at any layer for a model pretrained with dropout has lower loss than a model pretrained without, with lower loss for a model pretrained with higher dropout. Fig.~\ref{fig:early-exit-comparison-detailed:max_dropout_0.2} shows that uniform and ILD exhibit early exit improvements, but ALD does not.

Does a decreasing schedule sacrifice depth robustness by ending training dropout-free? Fig.~\ref{fig:early-exit-comparison-detailed:max_dropout_0.2} says no: models trained with decreasing schedules significantly outperform zero-dropout baselines at early exit, and Fig.~\ref{fig:early-exit-comparison} shows they match - and in some cases exceed - constant schedules when controlling for training FLOPs — proving that exposure to dropout during early training leaves lasting benefits.

\takeaway{Pre-training with layer dropout not only saves training FLOPs but always leads to better early exit loss during inference.}

\finding{For the same pre-training FLOPs budget, decreasing dropout schedule leads to the lowest validation loss as well as competitive early exit loss.}

Here, we have covered static early exit where all tokens exit at the same layer. We hypothesize that ``dynamic early exit'' where each token exits at a different layer based on a heurestic, router, or an auxiliary model (e.g., ~\citet{CALM}), will lead to better accuracy-throughput tradeoffs on a model pretrained with layer dropout. However, we leave verifying this hypothesis for future work.


\subsubsection{Intermediate Layer Skipping}

Layer dropout induces structural robustness enabling models to function when layers are skipped at inference. As shown in Fig.~\ref{fig:skip-layer-comparison:ald_zoom_out}, dense baselines exhibit immediate loss spikes when layers are skipped, whereas ALD facilitates graceful degradation. This zero-shot ``elastic'' effect allows a 906M model to bridge the gap toward smaller dense baselines, as shown in Fig.~\ref{fig:skip-layer-comparison:ald}, providing flexibility typically requiring complex modifications and/or continual pretraining, like LlamaFlex~\citep{llamaflex} or Flextron~\citep{flextron}, ``for free'' within the standard pre-training recipe.

\begin{figure}
    \centering
    \includegraphics[width=0.75\columnwidth]{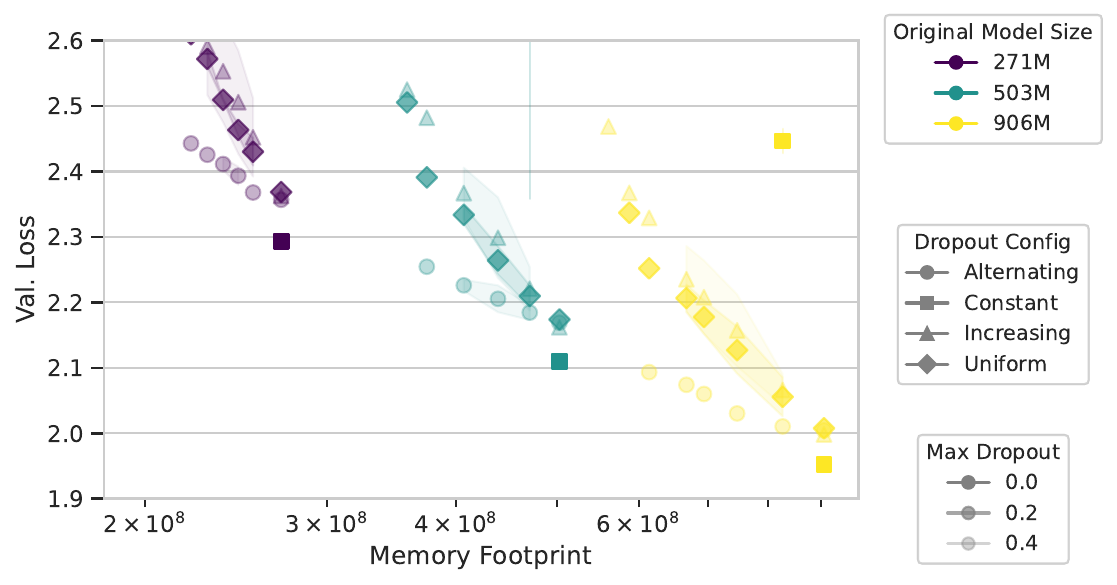}
    \caption{Intermediate layer skipping loss for models at 20 TPP: baseline vs. dropout configurations with 20\% FLOPs savings. Extended results in Fig.~\ref{fig:skip-layer-comparison-detailed}.}
    \label{fig:skip-layer-comparison:isoflops_reduction}
\end{figure}

Ablations at $p_{\text{max}}=0.2$ show that while all dropout variants improve skip-robustness, ALD offers superior retention for non-contiguous skipping (Fig.~\ref{fig:skip-layer-comparison:max_dropout_0.2}). Under iso-FLOP conditions (Fig.~\ref{fig:skip-layer-comparison:isoflops_reduction}), ALD maintains lower validation loss than ILD for an equivalent 20\% training compute reduction, confirming ALD as optimal for depth-wise inference elasticity at a fixed budget.

A clear trade-off emerges: ALD excels at skip-robustness but fails at early-exit, while ILD achieves better base accuracy and early-exit robustness with sub-optimal skip capability. Practitioners should choose based on deployment needs, and consider ALD with extended training to recover baseline accuracy.


\takeaway{Layer dropout induces an inherent elasticity that allows a larger model to gracefully degrade to the performance levels of smaller models when layers are skipped, providing a unified architecture for varying compute constraints.}

\finding{Average training dropout rate predicts zero-shot robustness to early exit and layer skipping without retraining.}


\subsection{Post-Training Inference Benefits}\label{sec:inference_optimizations:post-training}
While zero-shot techniques exploit the inherent redundancy of a model, further efficiency gains can be achieved through targeted post-training modifications that do not alter the pre-trained weights. We define post-training benefits as those derived from secondary training phases—such as continual pre-training or fine-tuning—specifically focused on optimizing inference throughput. In this work, we limit our investigation to ``weight-frozen'' methods where the transformer backbone remains static, and optimization is achieved by training auxiliary modules like adapters or routers. This paradigm ensures that the model’s foundational knowledge is preserved while expanding the Pareto-optimal frontier of its depth-wise flexibility. Here we cover using adapters and self-speculative decoding, and leave using routers (such as in~\citet{d-llm}) for future work.

\subsubsection{Early Exit Adapters}\label{sec:inference_optimizations:post-training:exit-adapter}

\paragraph{Background} 
To evaluate if the structural benefits of layer dropout persist after supervised optimization, we utilize the \emph{Balcony} framework for depth-based dynamic inference \citep{balcony}. Balcony is a lightweight approach that freezes the pre-trained backbone and inserts additional transformer layers as ``exit adapters'' at selected points. These adapters are trained using a self-distillation objective where a Kullback–Leibler (KL) divergence loss aligns intermediate sub-model outputs with the final layer's predictions. While \citet{balcony} demonstrates that incorporating these adapters directly into the pre-training phase leads to even lower early-exit loss compared to post-training addition, such joint training increases the memory footprint and FLOPs per step, potentially slowing down the pre-training process.

\begin{figure}[H]
    \centering
    \begin{subfigure}{0.32\textwidth}
        \centering
        \includegraphics[width=\linewidth]{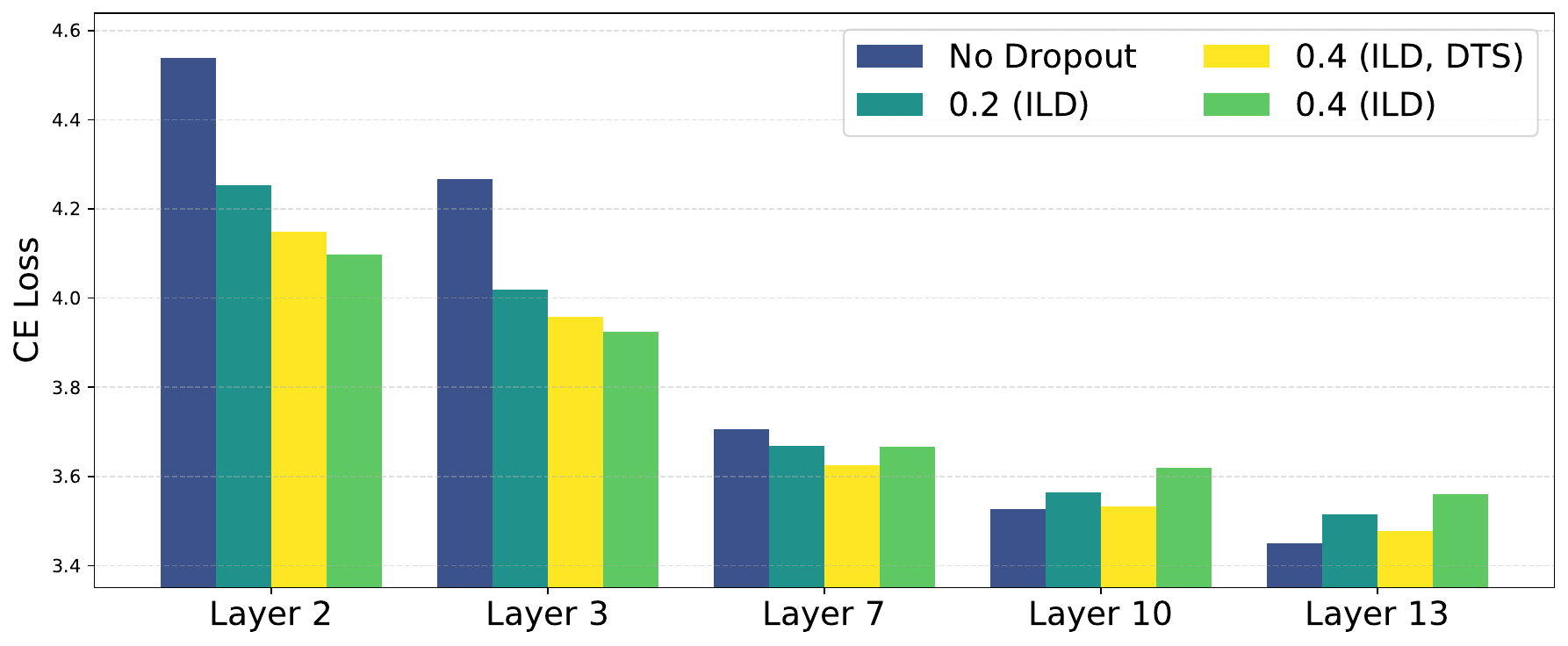}
        \caption{270M ($L=13$) 20 TPP.}
        \label{fig:balcony_20TPP:270m}
    \end{subfigure}
    \hfill
    \begin{subfigure}{0.32\textwidth}
        \centering
        \includegraphics[width=\linewidth]{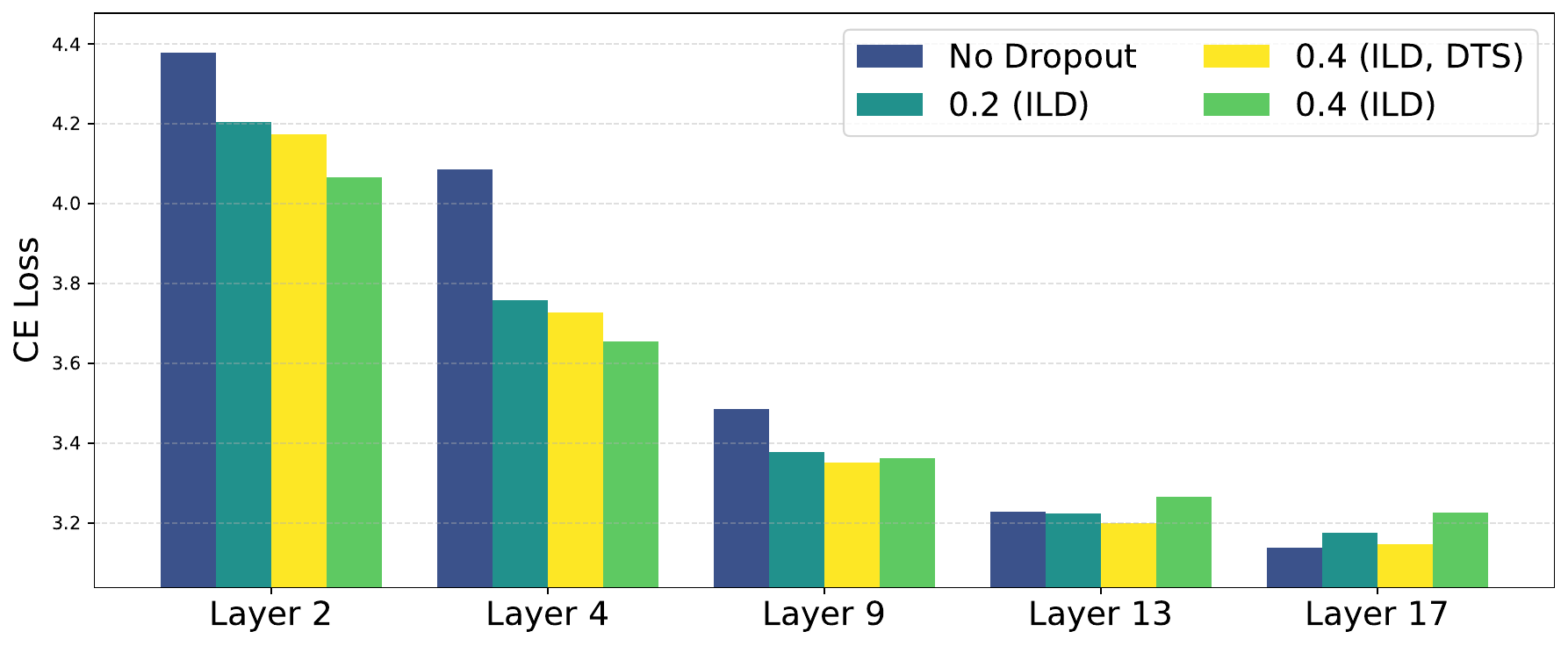}
        \caption{503M ($L=17$) 20 TPP.}
        \label{fig:balcony_20TPP:503m}
    \end{subfigure}
    \hfill
    \begin{subfigure}{0.32\textwidth}
        \centering
        \includegraphics[width=\linewidth]{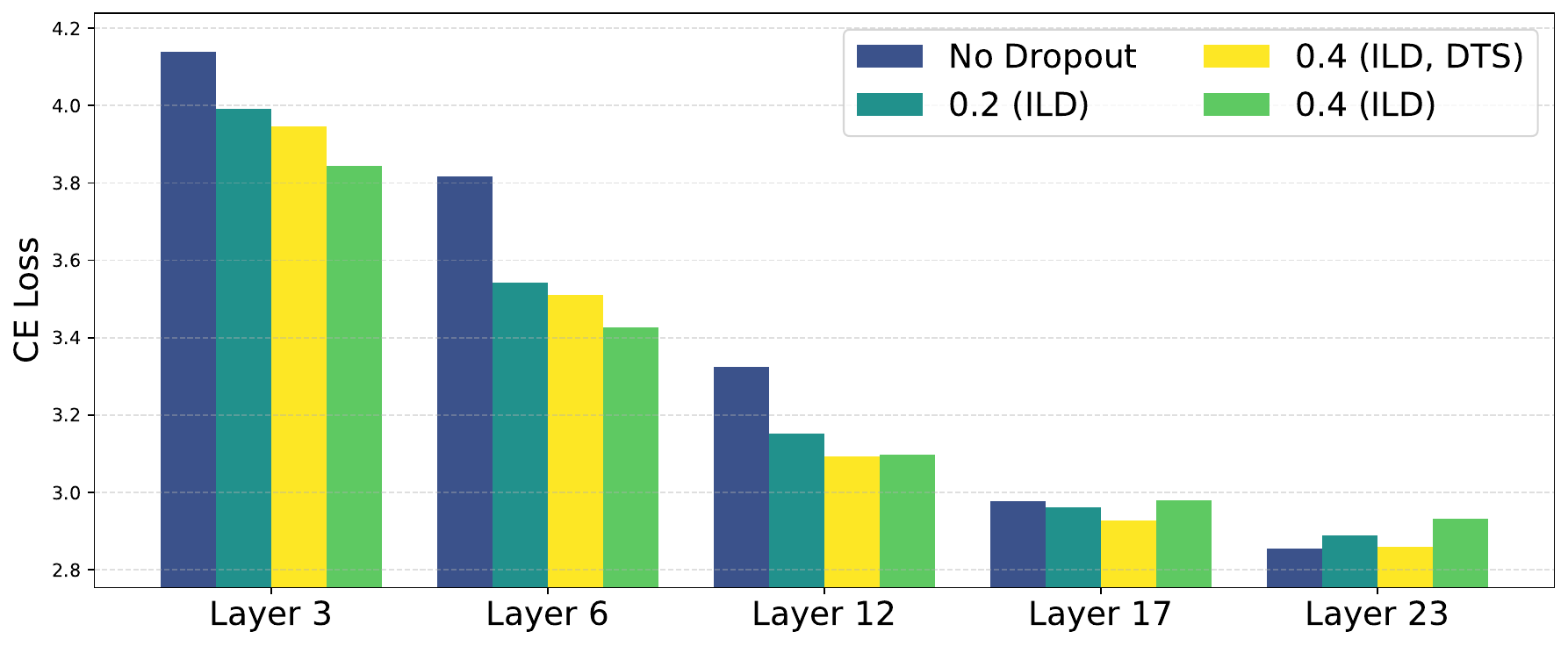}
        \caption{906M ($L=23$) 20 TPP.}
        \label{fig:balcony_20TPP:906m}
    \end{subfigure}

    \caption{Early exit losses for models pre-trained with different dropout configurations, followed by freezing their weights and training early exit adapters as proposed by Balcony~\citep{balcony}. Models pre-trained with dropout always lead to better early exit losses even after adding exit adapters.}
    \label{fig:balcony_20TPP}
\end{figure}

\paragraph{Analysis} 
As shown in Fig.~\ref{fig:balcony_20TPP}, models pre-trained with layer dropout consistently outperform dense baselines across all model scales—270M, 503M, and 906M—even when exit adapters are only added post-training. Our proposed method offers a more efficient alternative to joint adapter pre-training: by incorporating layer dropout, we improve the loss of earlier layers without increasing the memory footprint or computational overhead during pretraining. In fact, layer dropout actively reduces training FLOPs while inducing a permanent structural robustness that auxiliary training can leverage but cannot fully replicate on a standard dense model. Furthermore, models trained with higher dropout rates demonstrate a superior ``head-start'' for adapter training, reaching lower validation losses at earlier layers than their low-dropout counterparts, confirming that depth-aware pre-training is a prerequisite for maximizing the efficacy of post-training strategies.

\subsubsection{Self-Speculative Decoding}\label{sec:inference_optimizations:post-training:self-spec}

\paragraph{Background} Speculative decoding accelerates autoregressive inference by using a fast ``draft'' model to predict tokens that are validated in parallel by a larger ``target'' model, enabling lossless speedup \citep{speculative_decoding}. Self-speculative methods, such as \emph{Draft \& Verify}, use a subset of the target model's own layers to act as the drafter \citep{draft-and-verify}. The effectiveness of this approach depends on identifying a subset of layers that is small enough for high throughput yet accurate enough to maintain high token acceptance rates. 

\paragraph{Analysis} We hypothesized that the structural elasticity induced by layer dropout enables the discovery of more efficient subsets. While \citet{draft-and-verify} used Bayesian optimization to find draft layers, we also apply other search methods: genetic algorithms, hill climbing, and simulated annealing, and select the search result that leads to highest speedup. Results in Table~\ref{tab:speculative_decoding_speedup} confirm this showing models pre-trained with higher layer dropout obtain higher speedups during self-speculative decoding. For such models, search methods are able to find a subset of layers that achieve better trade offs of acceptance rate, $\alpha$, and draft decoding time, $T_{\text{Draft}}$. We leave for future work the evaluation of other self-speculative techniques like LayerSkip \citep{elhoushi-etal-2024-layerskip}, which uses early exit for drafting, and Kangaroo \citep{kangaroo}, which employs early exit with adapters trained in a similar manner to Balcony \citep{balcony}. 


\begin{table}[H]
\centering
\small
\setlength{\tabcolsep}{6pt}
\renewcommand{\arraystretch}{1.2}

\begin{tabular}{l c l c c c}
\toprule
\makecell[c]{\textbf{Model}\\\textbf{Size}} & \textbf{TPP} & \makecell[c]{\textbf{Dropout}\\\textbf{Config.}} & \multicolumn{3}{c}{\textbf{Self-Speculative Decoding}} \\
\cmidrule(lr){4-6}
& & & \makecell[c]{\textbf{Speedup}} & $\alpha$ & $\frac{T_{\text{Draft}}}{T_{\text{Target}}}$ \\
\midrule
\multirow{4}{*}{270M} & \multirow{4}{*}{20} & 0                  & 1.01$\times$ & 92\% & 0.65 \\
                      &                     & 0.2 (ILD)          & 1.20$\times$ & 86\% & 0.55 \\
                      &                     & 0.4 (ILD)          & \textbf{1.35$\times$} & 89\% & 0.48 \\
                      &                     & 0.4 (ILD, DTS)     & 1.22$\times$ & 73\% & 0.57 \\
\midrule
\multirow{4}{*}{504M} & \multirow{4}{*}{20} & 0                  & 1.03$\times$ & 90\% & 0.66 \\
                      &                     & 0.2 (ILD)          & 1.22$\times$ & 89\% & 0.53 \\
                      &                     & 0.4 (ILD)          & \textbf{1.43$\times$} & 97\% & 0.42 \\
                      &                     & 0.4 (ILD, DTS)     & 1.20$\times$ & 90\% & 0.52 \\
\midrule
\multirow{4}{*}{906M} & \multirow{4}{*}{20} & 0                  & 1.06$\times$ & 94\% & 0.66 \\
                      &                     & 0.2 (ILD)          & 1.32$\times$ & 98\% & 0.49 \\
                      &                     & 0.4 (ILD)          & \textbf{1.40$\times$} & 97\% & 0.48 \\
                      &                     & 0.4 (ILD, DTS)     & 1.29$\times$ & 95\% & 0.52 \\
\bottomrule
\end{tabular}

\caption{Self-speculative decoding speedup across model scales using Draft \& Verify~\citep{draft-and-verify} on XSUM~\citep{xsum}. $\alpha \in [0, 1]$ is acceptance rate for draft length $\gamma=5$, $T_\text{Draft}$ is time to decode a single token for the selected subset of layers, and $T_\text{Target}$ for the full model.}
\label{tab:speculative_decoding_speedup}

\end{table}

\section{Scaling Analysis}\label{sec:training_scaling}


In the preceding sections, we demonstrated that for compute-optimal pre-training at 20 TPP, specific layer dropout configurations—notably ILD+DTS—not only minimize degradation but can actually surpass the dense baseline in terms of validation accuracy. However, many ideas in architecture or pretraining in literature may perform well on small scale but fail to generalize at large scale, rendering them impractical to train foundational models. Hence, a critical question for modern foundational models is whether such benefits persist as the model is trained far beyond the compute-optimal point. 

Inspired by the predictable scaling frameworks established for model size and data budget \citep{hestness2017deeplearningscalingpredictable, kaplan2020scalinglawsneurallanguage, chinchilla}, and the recent investigation into how precision interacts with data scale \citep{kumar2024scalinglawsprecision}, we extend our analysis to high-TPP regimes. Our goal is to develop a scaling analysis that quantifies how the regularization and structural benefits of layer dropout evolve as the model exhausts its inherent redundancy through prolonged training. This allows us to predict the performance of sparse-trained models at the trillion-token scale typical of state-of-the-art LLMs.

\begin{figure}[h]
    \centering
    \includegraphics[width=0.5\columnwidth]{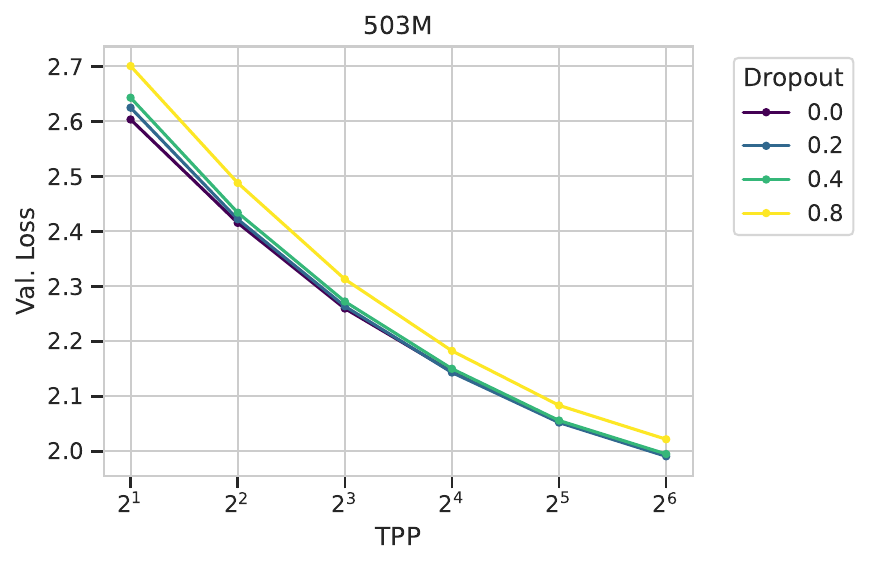}
    \caption{Validation loss across TPP for ILD with DTS, showing competitive performance with dense baselines as tokens-per-parameter increase. }
    \label{fig:training_scaling_tpp_reverse_schedule}
\end{figure}

Our recommended ILD+DTS configuration demonstrates that loss degradation remains remarkably stable as shown in Fig.~\ref{fig:training_scaling_tpp_reverse_schedule}, typically staying within $\approx$ 0.50\% of the baseline even at high TPP. 
These results indicate that layer dropout does not impede scaling performance. Instead, the stability observed suggests a robust, compute-efficient pre-training pathway that maintains structural benefits as we scale to training budgets of typical modern foundational LLMs.

\section{Large-Scale Runs}\label{sec:hero_run}

To conclude the empirical evaluation, this section presents scaling of our optimized layer dropout recipe to models exceeding the 1B parameter threshold with aggressive dropout rates. This analysis serves to validate our primary hypotheses: that larger model architectures exhibit inherently higher robustness to structural sparsity and that aggressive dropout rates are the primary enabler for depth-wise inference flexibility.

\begin{table}[h]
\centering
\caption{Training larger models with aggressive dropout preserves accuracy while accelerating training and inference.}
\small
\setlength{\tabcolsep}{3pt}
\begin{tabular}{l | c c | c c | c}
\toprule
\textbf{Model Size} & \multicolumn{2}{c|}{\textbf{1.8B}} & \multicolumn{2}{c|}{\textbf{3.9B}} & \multicolumn{1}{c}{\textbf{8.2B}} \\
\midrule
TPP & 20 & 20 & 20 & 20 & 20 \\
Max Dropout & 0 & 0.6 & 0 & 0.8 & 0.99 \\
Layer Dist. & -- & Inc. & -- & Inc. & Inc. \\
Time Sched. & -- & Dec. & -- & Dec. & Dec. \\
FLOPs Savings $\uparrow$ & 0\% & \textbf{15\%} & 0\% & \textbf{20\%} & \textbf{25\%} \\
\midrule
Val. Loss $\downarrow$ & 1.849 & \textbf{1.836} & \textbf{1.732} & 1.745 & 1.663 \\
Skip Alt. Layers $\downarrow$ & 4.260 & \textbf{2.282} & 6.446 & \textbf{2.129} & \textbf{1.991} \\
Early Exit @ $0.75L$ $\downarrow$ & 3.943 & \textbf{2.329} & 3.834 & \textbf{2.143} & \textbf{1.777} \\
Spec. Decode Speedup $\uparrow$ & 1.10$\times$ & \textbf{1.34$\times$} & 1.02$\times$ & \textbf{1.54$\times$} & \textbf{1.55$\times$} \\
\bottomrule
\end{tabular}
\label{tab:hero_run:efficiency}
\end{table}
\begin{figure*}
    \centering
    \begin{subfigure}[b]{0.4\textwidth}
        \centering
        \includegraphics[width=\linewidth]{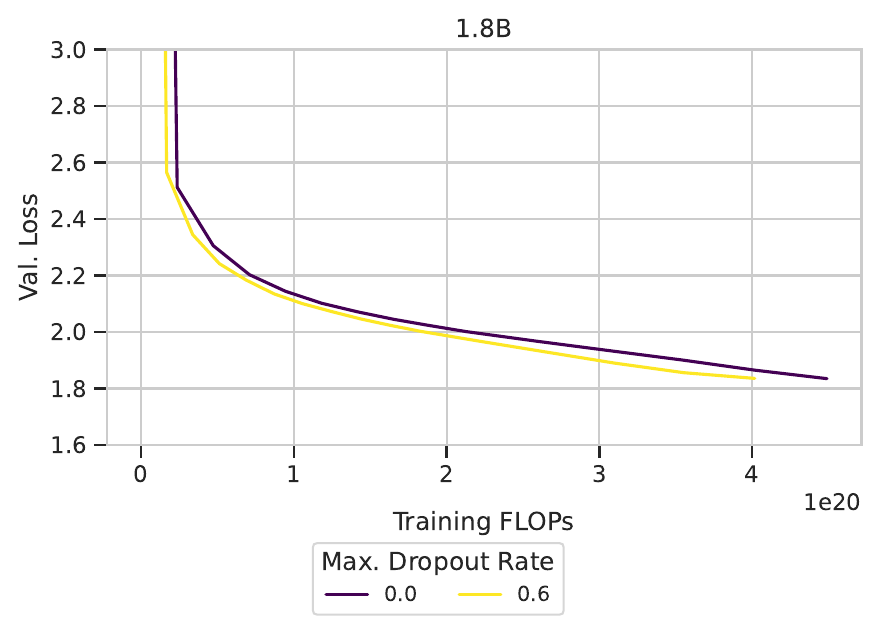}
        \label{fig:large_training_runs:1.8B}
    \end{subfigure}
    \
    \begin{subfigure}[b]{0.4\textwidth}
        \centering
        \includegraphics[width=\linewidth]{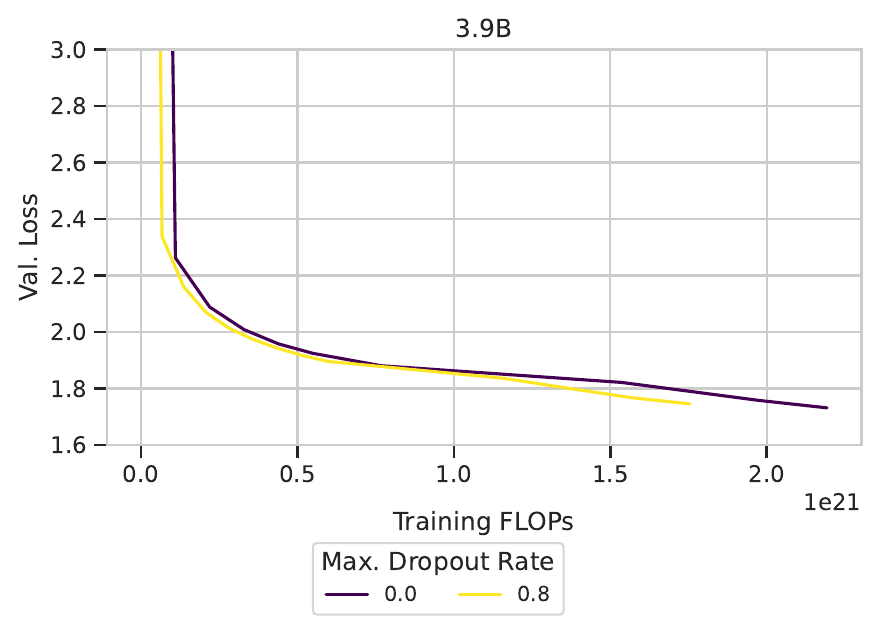}
        \label{fig:large_training_runs:3.9B}
    \end{subfigure}
    \caption{Validation loss versus training FLOPs for larger model sizes at aggressive maximum dropout rates for 20 TPP. At the same training FLOPs, training with layer dropout leads to better loss.}
    \label{fig:large_training_runs}
\end{figure*}

\paragraph{Pushing the Limits of Structural Sparsity}
While previous studies often limited dropout rates to conservative values, such as 0.1 or 0.2 \citep{vaswani2023attentionneed, gpt2}, we subject our 1.8B, 3.9B, 8.2B models to aggressive regimes with $p_{\text{max}}$ values of 0.6, 0.8, and 0.99, respectively. Taking 3.9B model as an example, the training initialization is significant: the effective depth of the model begins at only $0.6L$\footnote{Effective depth of a model with $L$ layers at iteration $t$ for ILD+DTS configuration is $\sum_{\ell=0}^{L-1}(1-p_{max}\frac{\ell}{L-1})=\frac{1+1-p_{max}}{2}L=(1-0.5p_{max})L$.}, with the final layer being skipped 80\% of the time. Despite this substantial reduction in early-training active capacity, the models converge to results that are competitive with, or superior to, the dense baselines in both validation loss (as shown in Table~\ref{tab:hero_run:efficiency}) and downstream task performance (as shown in Table~\ref{tab:all_downstream_tasks}). 

Furthermore, Figure~\ref{fig:large_training_runs} shows that training with dropout is not only faster, but also achieves lower loss for a given FLOP budget throughout most of training under aggressive maximum dropout rates. Based on trends across model sizes, we project that larger models will exhibit even greater robustness to high dropout rates, yielding increasingly pronounced validation loss improvements over dense baselines as scale increases.


\paragraph{Unlocking Inference Efficiency}
The operational advantages of this depth-aware pre-training are most evident in the inference benchmarks in Table~\ref{tab:hero_run:efficiency}. The 3.9B model exhibits a 1.54$\times$ speedup in self-speculative decoding, a task where the dense baseline fails to provide significant gain, resulting in a 1.02$\times$ regression. This confirms that the elasticity induced by aggressive dropout is a prerequisite for effective speculative drafting at these scales. Furthermore, the model shows high resilience to static pruning; skipping alternate layers results in a cross-entropy loss of 2.129 for the dropout-trained model, whereas the dense baseline's loss increases to 6.446. Figures \ref{fig:early-exit-comparison-detailed:large} and \ref{fig:skip-layer-comparison:large} show detailed results for skipping intermediate layers and early exit of those models.


\takeaway{At the multi-billion parameter scale, aggressive layer dropout ($p_{\text{max}}=0.99$) facilitates up to 25\% savings in cumulative training FLOPs while largely preserving baseline generalization performance, unlocking zero-shot inference speedups of up to 1.55$\times$.}

\section{Limitations}\label{sec:limitations}
While our results establish a robust framework for layer dropout at scale, our work has multiple limitations:
\begin{itemize}
    \item \textbf{Hyperparameter Transfer:} In our plots in Fig.~\ref{fig:hyperparameter_transfer_test}, although optimal learning rate, $\eta$, and weight decay, $\lambda$, remained largely the same for small to medium dropouts, they reduced for aggressively high dropouts. Improving transferrability to such high dropouts would improve our accuracy results further. Our transfer analysis was primarily validated for constant dropout schedules. Extending these rules to the decreasing schedules identified as optimal could potentially yield further accuracy improvements at high TPP budgets.
    \item \textbf{Alternative Granularities:} We focused exclusively on transformer-level dropout to induce depth-wise robustness. Whether other granularities, such as attention-head or neuron-level dropout, can induce similar structural resilience remains an open question.
    \item \textbf{Comparison with Learned Depth Optimization:} This study did not compare structured layer dropout against learned depth-aware mechanisms such as Mixture-of-Depths~\citep{MoD}. Investigating the trade-offs between stochastic layer removal and dynamic, routing-based depth optimization is a compelling direction for future work.
    \item \textbf{Cross-Architecture Generalization:} While we validated results up to 8.2B parameters, the interaction between aggressive layer dropout and alternative architectures, such as Mixture-of-Experts (MoE) or non-transformer models, has not yet been explored.
    \item \textbf{Scaling Laws for Maximum Dropout:} We have not yet developed comprehensive scaling laws to predict the maximum dropout rate ($p_{\text{max}}$) that can be applied without incurring accuracy degradation relative to the dense baseline.
    \item \textbf{Scaling Analysis for Inference Benefits:} We have not quantified the different inference benefits (early exit loss, skipping intermediate layer loss, early exit adapter loss, and self-speculative decoding speedup) as TPP increases.
\end{itemize}

\section{Conclusion}

In this study, we have demonstrated that layer dropout is not only a viable technique for the modern large-scale pre-training regime but a powerful mechanism for architectural flexibility. By systematically exploring dropout distributions and schedules, we have shown that a depth-aware training approach—specifically utilizing an increasing distribution across layers coupled with a decreasing schedule over time—can maintain, and in some cases surpass, dense baseline accuracy while significantly reducing training FLOPs, and unlocking zero-shot inference elasticity and speedups. This approach is non-invasive and orthogonal to existing architectural and training optimizations, making it easily adoptable in standard LLM pre-training stacks.

Our ``Hero Runs'' on 1.8B, 3.9B, and 8.2B models reveal that as architectures grow larger, their natural resilience to aggressive dropout rates grow, with our recommended recipe unlocking zero-shot inference speedups of up to 1.55$\times$. This inherent elasticity bridges the performance gap between discrete model sizes and enables high-efficiency deployment strategies, such as self-speculative decoding, that may fail on standard dense models.


Our findings suggest a generalized training curriculum: progressively increasing a model's effective capacity throughout training yields superior results. While this work focused on increasing effective depth via layer dropout, this ``model growing'' strategy can be extended to other dimensions—such as model width—and granularities—such as quantization bit-widths, or unstructured sparsity—using similar spatial distributions and temporal schedules. We envision this framework as a foundational pillar for efficient large-scale pre-training, encouraging the re-adoption of layer dropout as a primary enabler for flexible model growth.

Future work can also include deducing the optimal maximum dropout rates for specific model scales and data budgets to maximize the Pareto frontier of training and inference efficiency, as well as investigating ``learned'' depth-aware training mechanisms, where the model dynamically identifies optimal skipping paths rather than relying on stochastic selection. 

\section*{Impact Statement}
This paper presents work whose goal is to advance the field of machine learning. There are many potential societal consequences of our work, none of which we feel must be specifically highlighted here.

\section*{Acknowledgments}

We would like to thank Shaheer Mohammad and Sam McPhail for infrastructure support at Cerebras.

\bibliographystyle{cereb}
\bibliography{references}

\newpage
\appendix
\onecolumn

\renewcommand{\thefigure}{A.\arabic{figure}}
\renewcommand{\thetable}{A.\arabic{table}}
\setcounter{figure}{0}
\setcounter{table}{0}

{\Large\bfseries Appendix}
\vspace{1em}

\newglossaryentry{FFN} {
    name=FFN,
    description={Feed Forward Networks}
}

\newglossaryentry{tpp}{
    name=TPP,
    description={Tokens per Parameter}
}

\newglossaryentry{layer_dropout} {
    name=Layer Dropout,
    description={A form of dropout where entire layers of a neural network (e.g., transformer blocks) are randomly skipped during training. Unlike standard dropout, which zeroes out individual activations, layer dropout operates at the structural level, reducing the effective depth of the network on each forward pass. In the context of transformers, we use this term to indicate applying layer dropout on the granularity of a whole transformer block}
}

\newglossaryentry{stochastic_depth}{
    name=Stochastic Depth,
    description={An alternative term for \Gls{layer_dropout}}
}

\newglossaryentry{sub_layer_dropout} {
    name=Sub-Layer Dropout,
    description={In the context of transformers, refers to applying layer dropout on attention and FFN blocks independently}
}

\newglossaryentry{uniform_distribution}{
    name=Uniform Distribution,
    description={A layer dropout configuration where dropout rates of all layers are set to the same value, and can be mathematically expressed as $p^{\ell,t}_{\text{uniform}}=p_{\text{max}}$}
}

\newglossaryentry{increasing_layer_distribution}{
    name=Increasing Layer Distribution (ILD),
    description={A layer dropout configuration where the dropout rate starts at 0 for the first layer, i.e., $p^{\ell=0}=0$, and linearly increases across layers to a maximum dropout rate. It is mathematically expressed as: $p^{\ell,t}_{\text{ILD}} = \frac{\ell}{L-1} \cdot p_{\text{max}}$}
}

\newglossaryentry{alternating_layer_distribution}{
    name=Alternating Layer Distribution (ALD),
    description={A dropout distribution where dropout is not applied on the first layer, but is applied on alternating layers proceeding that. It is expressed mathematically as: $p^{\ell,t}_{\text{ALD}} = p_{\text{max}} \cdot \mathbf{1}_{\ell\equiv 1 \text{(mod 2)}}$}
}

\newglossaryentry{constant_time_schedule}{
    name=Constant Time Schedule,
    description={A temporal schedule for layer dropout where the dropout rate remains constant throughout training. It is expressed mathematically as $p^{\ell,t}_{\text{const}} = p^{\ell}_{\text{dist}}$}
}

\newglossaryentry{increasing_time_schedule}{
    name=Increasing Time Schedule (ITS),
    description={A temporal schedule for layer dropout where the dropout rate starts at zero and increases linearly to its maximum by the final training step. It is expressed mathematically as $p^{\ell,t}_{\text{ITS}} = p^{\ell}_{\text{dist}} \cdot \left( \frac{t}{T} \right)$}
}

\newglossaryentry{decreasing_time_schedule}{
    name=Decreasing Time Schedule (DTS),
    description={A temporal schedule for layer dropout where the dropout rate is at its maximum at the start of pre-training and decays linearly to zero by the final training step. It is expressed mathematically as $p^{\ell,t}_{\text{DTS}} = p^{\ell}_{\text{dist}} \cdot \left( 1 - \frac{t}{T} \right)$. In this work, we demonstrate that this schedule helps stabilize early training while allowing the model to settle into a dense state for final convergence}
}

\printglossaries
\newpage

\section{Experimental Settings}

\begin{table}[H]
\centering
\small
\setlength{\tabcolsep}{4pt}
\begin{tabular}{lccccc}
\hline
\textbf{Model} &
\makecell{\textbf{Hidden Dim.}\\$D$} &
\makecell{\textbf{Layers}\\$L$} &
\makecell{\textbf{Heads}\\} &
\makecell{\textbf{Head Size}\\} &
\makecell{\textbf{FFN Mult.}\\} \\
\hline
271M  & 640  & 13 & 10 & 64  & 8 \\
504M  & 896  & 17 & 14 & 64  & 8 \\
906M  & 1152 & 23 & 9  & 128 & 8 \\
1.8B  & 1536 & 30 & 12 & 128 & 8 \\
3.9B  & 2048 & 40 & 16 & 128 & 8 \\
8.2B  & 2688 & 52 & 21 & 128 & 8 \\
\hline
\end{tabular}
\caption{Model architectures used in the experiments.}
\label{tab:model-configs}
\end{table}

\begin{table}[H]
    \centering
    \vspace{-4pt}
    \caption{\small
        Summary of SP, \textmu P, and CompleteP with layer dropout rate for a transformer model. Terms related to \begingroup\color{orange}width \endgroup (introduced by $\mu$P~\citet{mup}), \begingroup\color{ForestGreen}depth \endgroup (introduced by CompleteP~\citet{CompleteP}), \begingroup\color{brown}data size \endgroup (introduced by ~\citet{PowerLines}), and \begingroup\color{purple}dropout \endgroup (introduced in this paper) controls are highlighted in \begingroup\color{orange}orange\endgroup, \begingroup\color{ForestGreen}green\endgroup, \begingroup\color{brown}brown\endgroup, and  \begingroup\color{purple}purple \endgroup respectively. \begingroup\color{blue}Additional tunable parameters\endgroup\ are highlighted in \begingroup\color{blue}blue\endgroup. \emph{Hidden} refers to all linear layers in the transformer backbone. Layer density, \begingroup\color{purple}$\rho$\endgroup\ is the complement of layer dropout, \begingroup\color{purple}$p$\endgroup\ such that \begingroup\color{purple}$\rho=1-p$\endgroup\ .  
    }
    \label{tab:parameterization-summary}
    \resizebox{0.8\textwidth}{!}
    {
    \begin{tabular}{lll}
        \toprule
        Parameterization        & Base Training 
                                & Scaled Training \\
        \midrule
        Width                   & $d_{\text{base}}$
                                & $d_{\text{base}} \cdot \begingroup\color{orange}m_d\endgroup $\\
        Depth                   & $L_{\text{base}}$ 
                                & $L_{\text{base}} \cdot \begingroup\color{ForestGreen}{m_L}\endgroup $ \\
        Dataset Size            & $D_{\text{base}}$
                                & $D_{\text{base}} \cdot \begingroup\color{brown}{m_D}\endgroup $\\
        Layer Density           & $1$
                                & $\begingroup\color{purple}{\rho}\endgroup$ \\
        Model Size              & $N_{\text{base}} = g(d_{\text{base}}, L_{\text{base}})$
                                & $N = g(d_{\text{base}} \cdot \begingroup\color{orange}m_d\endgroup, L_{\text{base}} \cdot \begingroup\color{ForestGreen}{m_L}\endgroup)$\\
        Tokens per Parameter    & $\text{TPP}_{\text{base}} = D_{\text{base}} / N_{\text{base}} $
                                & $\text{TPP} = D_{\text{base}} \cdot \begingroup\color{brown}{m_D}\endgroup / N $ \\
        \midrule
        Batch Size              & $\begingroup\color{blue}\ B_{\text{base}}\endgroup$
                                & $\begingroup\color{blue}\ B_{\text{base}} \endgroup \cdot \begingroup\color{brown}{m_D^{0.4}}\endgroup $ \\
        Timescale (AdamW)       & $\begingroup\color{blue}\tau_{\text{EMA}_{\text{base}}} \endgroup $
                                & $ \tau_{\text{EMA}} = \begingroup\color{blue}\tau_{\text{EMA}_{\text{base}}} \endgroup \cdot \begingroup\color{brown}(\frac{\text{TPP}}{\text{TPP}_{\text{base}}})^{-0.5}\endgroup$ \\
        \midrule
        Emb. Init. Var.        & $\begingroup\color{blue}\sigma_{\text{base}}^2\endgroup$ 
                                & $\begingroup\color{blue}\sigma_{\text{base}}^2\endgroup$ \\
        Emb. LR (AdamW)        & $\begingroup\color{blue}\eta_{\text{base}}\endgroup$
                                & $\begingroup\color{blue}\eta_{\text{base}}\endgroup$\\ 
        \midrule
        Pre-LN Init. Var.       & $\begingroup\color{blue}\sigma_{\text{base}}^2\endgroup$
                                & $\begingroup\color{blue}\sigma_{\text{base}}^2\endgroup$ \\
        Pre-LN LR (AdamW)       & $\begingroup\color{blue}\eta_{\text{base}}\endgroup$
                                & $\begingroup\color{blue}\eta_{\text{base}}\endgroup\begingroup\color{ForestGreen}\endgroup$ \\ 
        \midrule
        Hidden Init. Var.       & $\begingroup\color{blue}\sigma_{\text{base}}^2\endgroup$
                                & $\begingroup\color{blue}\sigma_{\text{base}}^2\endgroup \cdot \begingroup\color{orange}m_d^{-1}\endgroup$ \\
        Hidden LR (AdamW)       & $\begingroup\color{blue}\eta_{\text{base}}\endgroup$
                                & $\begingroup\color{blue}\eta_{\text{base}}\endgroup  \cdot \begingroup\color{orange}m_d^{-1}\endgroup$ \\
        Hidden Bias LR (AdamW)  & $\begingroup\color{blue}\eta_{\text{base}}\endgroup$
                                & $\begingroup\color{blue}\eta_{\text{base}}\endgroup$ \\
        Hidden WD (AdamW)       & $\frac{\begingroup\color{blue}\ B_{\text{base}}\endgroup}{\begingroup\color{blue}\eta_{\text{base}} \tau_{\text{EMA}_{\text{base}}} \endgroup D_{\text{base}}}$ 
                                & $ \frac{
  \begingroup\color{blue} B_{\text{base}} \endgroup
}{
  \begingroup\color{blue} \eta_{\text{base}} \endgroup\,
  \tau_{\text{EMA}}\,
  D_{\text{base}}\,
  \begingroup\color{brown} m_D^{0.4} \endgroup
}
\cdot
\begingroup\color{orange} m_d \endgroup $ \\
        \midrule
        Attention Residual            & $\XB^l + f_{\text{attn}}(\XB^l)$
                                & $\XB^l + \begingroup\color{ForestGreen}{m_L}^{-1}\endgroup \begingroup\color{purple}{\rho}^{-1}\endgroup \cdot f_{\text{attn}}(\XB^l)$ \\
        FFN Residual            & $\ZB^l + f_{\text{ffn}}(\ZB^l)$
                                & $\ZB^l + \begingroup\color{ForestGreen}{m_L}^{-1}\endgroup \begingroup\color{purple}{\rho}^{-1}\endgroup \cdot f_{\text{ffn}}(\ZB^l)$ \\
        \midrule
        Final-LN Init. Var.     & $\begingroup\color{blue}\sigma_{\text{base}}^2\endgroup$
                                & $\begingroup\color{blue}\sigma_{\text{base}}^2\endgroup$ \\
        Final-LN LR (AdamW)     & $\begingroup\color{blue}\eta_{\text{base}}\endgroup$
                                & $\begingroup\color{blue}\eta_{\text{base}}\endgroup$ \\ 
        \midrule
        Unemb. Init. Var.      & $\begingroup\color{blue}\sigma_{\text{base}}^2\endgroup$ 
                                & $\begingroup\color{blue}\sigma_{\text{base}}^2\endgroup$ \\
        Unemb. LR (AdamW)      & $\begingroup\color{blue}\eta_{\text{base}}\endgroup$
                                & $\begingroup\color{blue}\eta_{\text{base}}\endgroup$\\ 
        Unemb. Fwd.            & $\XB^{L} \WB^\top_{\text{unemb}}$
                                & $\XB^{L} \WB^\top_{\text{unemb}} \cdot \begingroup\color{orange}m_d^{-1}\endgroup$ \\
        \midrule
        AdamW $\epsilon$ (Residual blocks)       & $\begingroup\color{blue}\epsilon_{\text{base}}\endgroup$
                                & $\begingroup\color{blue}\epsilon_{\text{base}}\endgroup \cdot \begingroup\color{orange}m_d^{-1}\endgroup \cdot \begingroup\color{ForestGreen}{m_L}^{-1}\endgroup$ \\
        AdamW $\epsilon$ (Emb. \& Unemb.)       & $\begingroup\color{blue}\epsilon_{\text{base}}\endgroup$
                                & $\begingroup\color{blue}\epsilon_{\text{base}}\endgroup \cdot \begingroup\color{orange}m_d^{-1}\endgroup$ \\
        \bottomrule
    \end{tabular}
    } %
\end{table}

\section{Coordinate Check}\label{sec:coordinate_check}

\begin{figure}[H]
  \centering
  \begin{subfigure}[t]{0.25\linewidth}
    \centering
    \includegraphics[width=\linewidth]{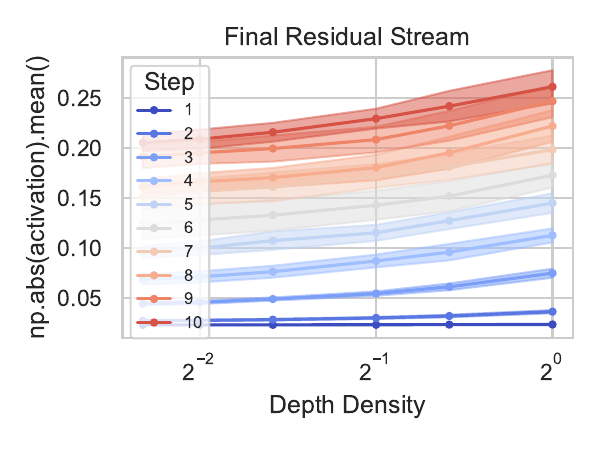}
    \caption{$r^{l,t}_{\text{train}}=1$}
    \label{fig:coordinate_check_nonuniform_distribution:one}
  \end{subfigure}
  \begin{subfigure}[t]{0.25\linewidth}
    \centering
    \includegraphics[width=\linewidth]{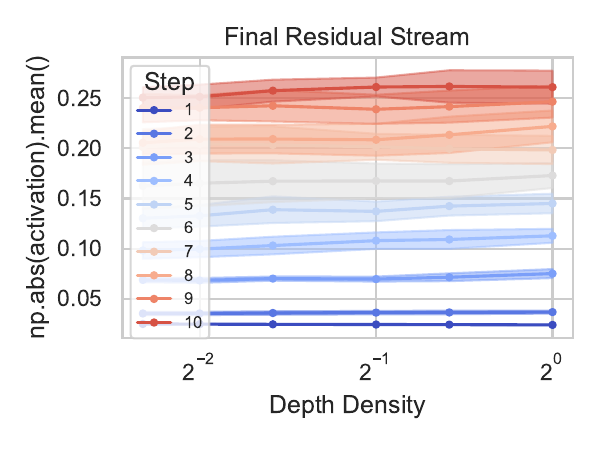}
    \caption{$r^{l,t}_{\text{train}}=1/\rho^{l,t}$ \crown}
    \label{fig:coordinate_check_nonuniform_distribution:one_over_rho}
  \end{subfigure}

    \caption{Coordinate Check passing for non-uniform distribution.} 
    \label{fig:coordinate_check_nonuniform_distribution}
    \vspace{-20pt}
\end{figure}

Fig.~\ref{fig:coordinate_checks_uniform_dropout} and Fig.~\ref{fig:coordinate_check_nonuniform_distribution} show coordinate check plots for uniform and non-uniform dropouts respectively. They show Frobenius norm of activations after merged residual streams from attention
and FFN blocks across a 40 layer model after 10 training steps, using CompleteP (that scales residuals by $\frac{L_{\text{base}}}{L}$), $r^{\ell}_{\text{train}}$ for layer $\ell$. 

\section{Additional Results}

\subsection{Downstream Tasks}
\begin{table}[H]
\centering
\caption{Downstream task performance benchmarks.}
\scriptsize
\setlength{\tabcolsep}{2pt}
\begin{tabular}{l c | l c | c c c c c c c c c c c}
\toprule
 &  & \multicolumn{2}{c|}{\textbf{Training Config.}} & \multicolumn{11}{c}{\textbf{Downstream Tasks $\uparrow$}} \\
\cmidrule(lr){3-4} \cmidrule(lr){5-15}
\textbf{Size} & \textbf{TPP} & \makecell{\textbf{Layer}\\\textbf{Dropout}} & \makecell{\textbf{FLOPs}\\\textbf{Sav.} $\uparrow$} 
& \textbf{BBH} & \textbf{PIQA} & \textbf{SIQA} & \textbf{Hella.} & \textbf{Wino.} & \textbf{ARC-c} & \textbf{ARC-e} & \textbf{OBQA} & \textbf{Lamb.} & \textbf{COPA} & \textbf{RACE} \\
\midrule

271M & 20 & 0 & 0\%
& 0.083 & 0.593 & 0.358 & 0.293 & 0.500 & 0.224 & 0.415 & 0.274 & 0.226 & 0.580 & 0.265 \\
271M & 20 & 0.2 (ILD) & 10\%
& 0.095 & 0.579 & 0.349 & 0.285 & 0.505 & 0.224 & 0.399 & 0.262 & 0.218 & 0.580 & 0.273 \\
271M & 20 & 0.4 (ILD, DTS) & 10\%
& 0.106 & 0.584 & 0.358 & 0.290 & 0.523 & 0.241 & 0.409 & 0.274 & 0.229 & 0.600 & 0.269 \\
271M & 20 & 0.4 (ILD) & 20\%
& 0.128 & 0.584 & 0.345 & 0.283 & 0.499 & 0.234 & 0.400 & 0.266 & 0.223 & 0.610 & 0.253 \\

\midrule

503M & 20 & 0 & 0\%
& 0.178 & 0.594 & 0.360 & 0.316 & 0.515 & 0.246 & 0.441 & 0.282 & 0.280 & 0.560 & 0.279 \\
503M & 20 & 0.2 (ILD) & 10\%
& 0.198 & 0.592 & 0.357 & 0.306 & 0.497 & 0.234 & 0.446 & 0.272 & 0.290 & 0.620 & 0.266 \\
503M & 20 & 0.4 (ILD, DTS) & 10\%
& 0.162 & 0.586 & 0.361 & 0.309 & 0.516 & 0.239 & 0.443 & 0.274 & 0.281 & 0.670 & 0.268 \\
503M & 20 & 0.4 (ILD) & 20\%
& 0.167 & 0.596 & 0.358 & 0.301 & 0.499 & 0.241 & 0.434 & 0.270 & 0.283 & 0.640 & 0.263 \\

\midrule

906M & 20 & 0 & 0\%
& 0.206 & 0.611 & 0.360 & 0.353 & 0.493 & 0.242 & 0.493 & 0.286 & 0.337 & 0.660 & 0.286 \\
906M & 20 & 0.2 (ILD) & 10\%
& 0.220 & 0.607 & 0.386 & 0.343 & 0.510 & 0.242 & 0.477 & 0.288 & 0.339 & 0.670 & 0.300 \\
906M & 20 & 0.4 (ILD, DTS) & 10\%
& 0.226 & 0.609 & 0.365 & 0.353 & 0.527 & 0.257 & 0.484 & 0.272 & 0.345 & 0.680 & 0.305 \\
906M & 20 & 0.4 (ILD) & 20\%
& 0.221 & 0.613 & 0.357 & 0.334 & 0.513 & 0.245 & 0.471 & 0.274 & 0.330 & 0.650 & 0.292 \\

\midrule

1.9B & 20 & 0 & 0\%
& 0.251 & 0.640 & 0.385 & 0.404 & 0.528 & 0.282 & 0.543 & 0.290 & 0.398 & 0.670 & 0.313 \\
1.9B & 20 & 0.6 (ILD, DTS) & 15\%
& 0.241 & 0.636 & 0.376 & 0.400 & 0.5241 & 0.282 & 0.533 & 0.282 & 0.389 & 0.670 & 0.299 \\

\midrule

3.9B & 20 & 0 & 0\%
& 0.272 & 0.666 & 0.399 & 0.473 & 0.555 & 0.312 & 0.597 & 0.356 & 0.479 & 0.650 & 0.325 \\
3.9B & 20 & 0.8 (ILD, DTS) & 20\%
& 0.270 & 0.668 & 0.402 & 0.462 & 0.563 & 0.317 & 0.585 & 0.324 & 0.480 & 0.680 & 0.329 \\

\midrule

8.2B & 20 & 0.99 (ILD, DTS) & 25\%
& 0.291 & 0.699 & 0.407 & 0.523 & 0.566 & 0.360 & 0.634 & 0.356 & 0.529 & 0.700 & 0.356 \\

\bottomrule
\end{tabular}
\label{tab:all_downstream_tasks}
\end{table}

\subsection{Zero-Shot Inference Benefits}

\subsubsection{Early Exit}
\begin{figure}[H]
    \centering

    \begin{subfigure}{\textwidth}
        \centering
        \includegraphics[width=\textwidth]{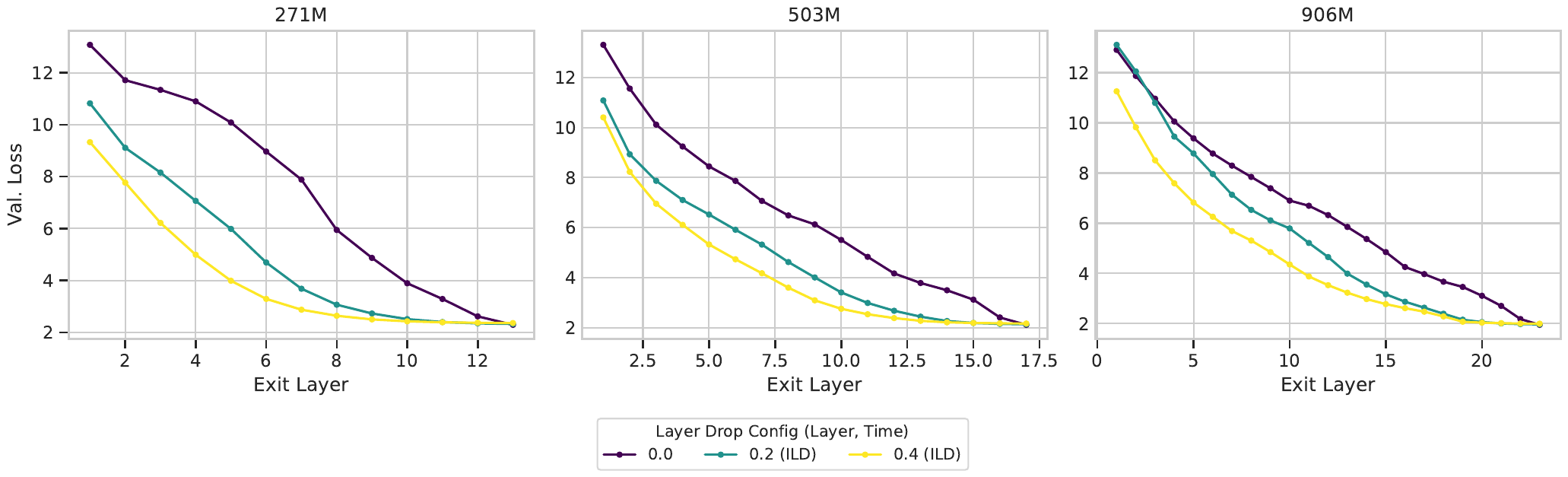}
        \caption{Ablating different maximum dropout values for Linear Distribution.}
        \label{fig:early-exit-comparison-detailed:pld_only}
    \end{subfigure}

    \vspace{1em}

    \begin{subfigure}{\textwidth}
        \centering
        \includegraphics[width=\textwidth]{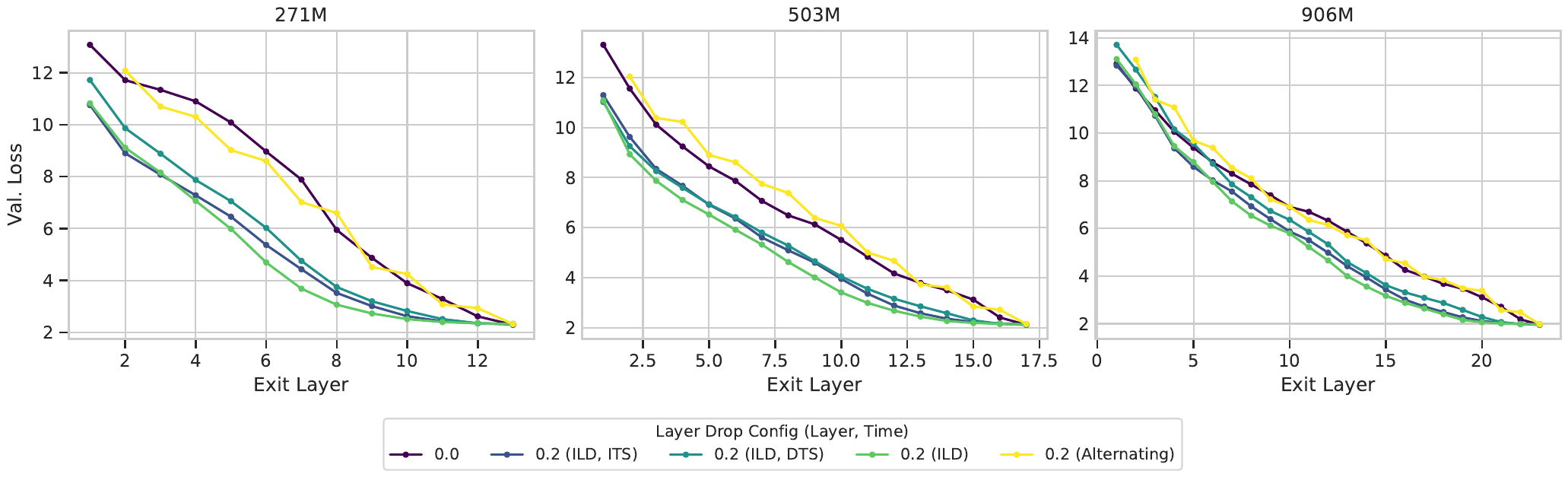}
        \caption{Ablating different dropout configurations for the same maximum dropout of 0.2.}
        \label{fig:early-exit-comparison-detailed:max_dropout_0.2}
    \end{subfigure}

    \vspace{1em}


    \begin{subfigure}{\textwidth}
        \centering
        \includegraphics[width=\textwidth]{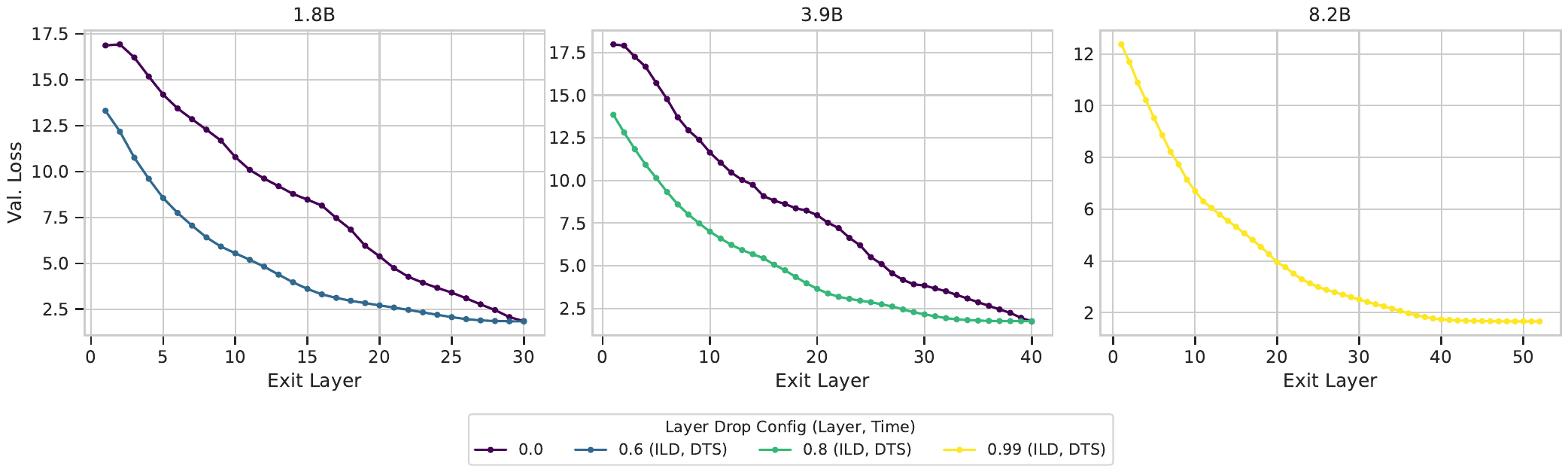}
        \caption{Larger model sizes with aggressive maximum dropout rates.}
        \label{fig:early-exit-comparison-detailed:large}
    \end{subfigure}

    \caption{Comparison of early-exit validation losses for models trained with different dropout configurations. All models trained with 20 TPP.}
    \label{fig:early-exit-comparison-detailed}
\end{figure}

\subsubsection{Intermediate Layer Skipping}
\begin{figure}[H]
    \centering

    \begin{subfigure}[t]{0.49\textwidth}
        \centering
        \includegraphics[width=\textwidth]{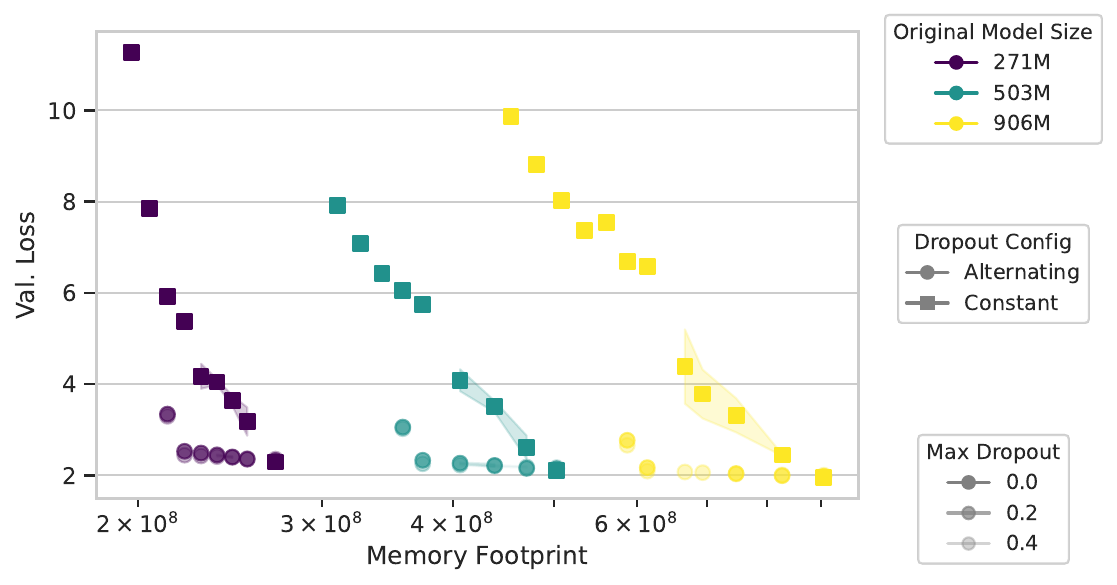}
        \caption{Ablating different maximum dropout values for Alternate Distribution.}
        \label{fig:skip-layer-comparison:ald_zoom_out}
    \end{subfigure}
    \hfill
    \begin{subfigure}[t]{0.49\textwidth}
        \centering
        \includegraphics[width=\textwidth]{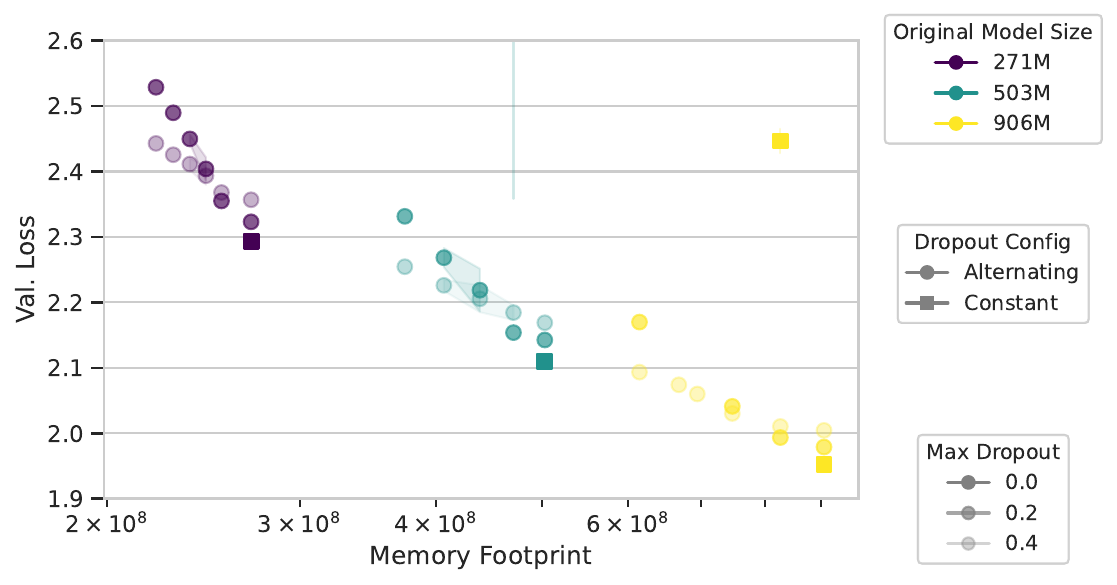}
        \caption{Ablating different maximum dropout values for Alternate Distribution, zooming on skipping configurations with lower losses.}
        \label{fig:skip-layer-comparison:ald}
    \end{subfigure}

    \vspace{1em}

    \begin{subfigure}[t]{0.49\textwidth}
        \centering
        \includegraphics[width=\textwidth]{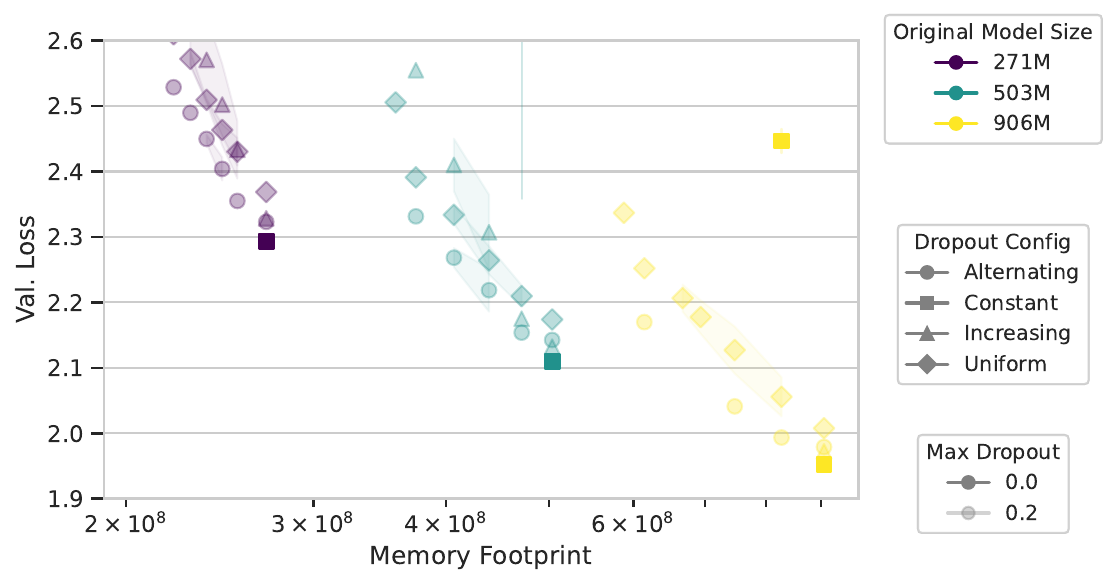}
        \caption{Ablating different dropout configurations for the same maximum dropout of 0.2.}
        \label{fig:skip-layer-comparison:max_dropout_0.2}
    \end{subfigure}
    \hfill
    \begin{subfigure}[t]{0.49\textwidth}
        \centering
        \includegraphics[width=\textwidth]{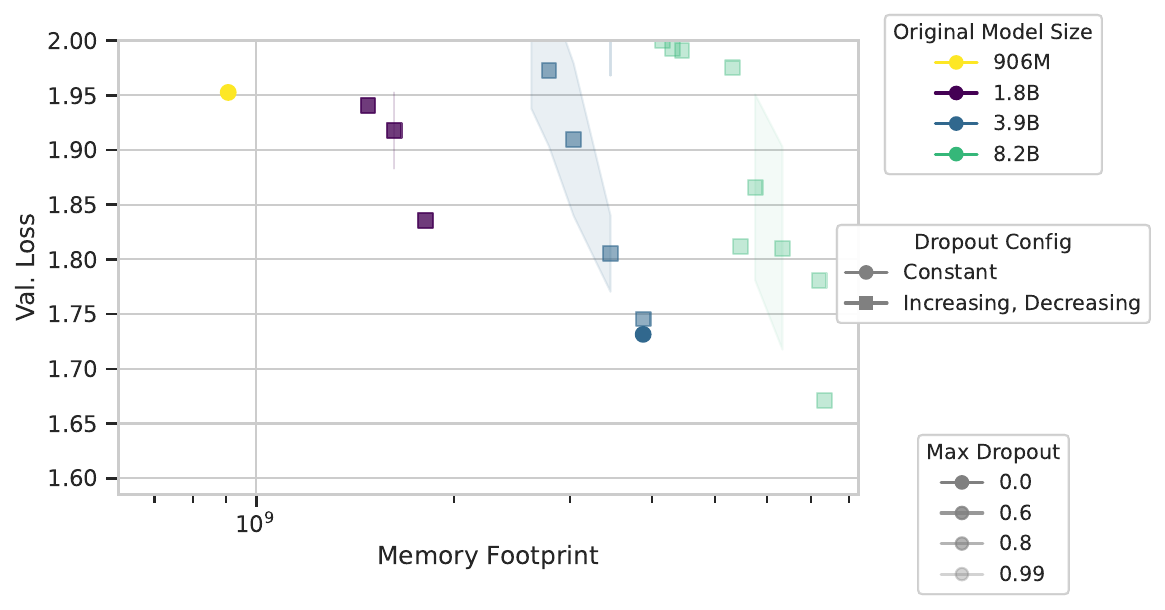}
        \caption{Larger model sizes with aggressive maximum dropout rates.}
        \label{fig:skip-layer-comparison:large}
    \end{subfigure}

    \caption{Comparison of intermediate layer skipping validation losses for models trained with different dropout configurations. All models trained with 20 TPP.}
    \label{fig:skip-layer-comparison-detailed}
\end{figure}

\end{document}